%% file: neurips_2025.tex
\documentclass{article}

    \PassOptionsToPackage{numbers, compress}{natbib}

\usepackage[final]{neurips_2025}

\usepackage[utf8]{inputenc} 
\usepackage[T1]{fontenc}    
\usepackage{hyperref}       
\usepackage{url}            
\usepackage{booktabs}       
\usepackage{amsfonts}       
\usepackage{nicefrac}       
\usepackage{microtype}      
\usepackage{xcolor}         
\usepackage{graphicx}

\usepackage{enumitem}
\usepackage{amsmath}
\usepackage{amssymb}
\usepackage{mathtools}
\usepackage{amsthm}
\usepackage{makecell}
\usepackage{multirow}
\usepackage{pifont}
\usepackage{wrapfig}
\usepackage{algorithm}
\usepackage{subcaption}
\usepackage{algpseudocode}
\usepackage[table,svgnames]{xcolor}
\definecolor{lightblue}{RGB}{239,247,255}

\theoremstyle{plain}

\theoremstyle{definition}

\theoremstyle{remark}

\newcommand{\cross}{\ding{55}}

\newcommand{\ie}{\emph{i.e., }}
\newcommand{\eg}{\emph{e.g., }}

\newcommand{\etc}{\emph{etc.}}

\newcommand{\yuan}[1]{{\color{black}{#1}}}
\newcommand{\wuc}[1]{{\color{black}{#1}}}

\newcommand{\liang}[1]{{\color{black}{#1}}}

\newcommand{\reb}[1]{{\color{black}{#1}}}

\usepackage[capitalize]{cleveref}
\crefname{figure}{Figure}{Figures}
\crefname{table}{Table}{Tables}

\title{3D-GSRD: 3D Molecular Graph Auto-Encoder \\ with Selective Re-mask Decoding}

\author{
\textbf{Chang Wu\textsuperscript{1*}},
\textbf{Zhiyuan Liu\textsuperscript{2*}},
\textbf{Wen Shu\textsuperscript{3}},
\textbf{Liang Wang\textsuperscript{4}},
\textbf{Yanchen Luo\textsuperscript{1}},\\
\textbf{Wenqiang Lei\textsuperscript{3}},
\textbf{Yatao Bian\textsuperscript{2}},
\textbf{Junfeng Fang\textsuperscript{2$\dagger$}},
\textbf{Xiang Wang\textsuperscript{1$\dagger$}} \\[3mm]
\textsuperscript{1} University of Science and Technology of China 
\textsuperscript{2} National University of Singapore \\
\textsuperscript{3} Sichuan University 
\textsuperscript{4} Institute of Automation, Chinese Academy of Sciences \\[2mm]
\small \texttt{wuchang0124@mail.ustc.edu.cn}, 
\texttt{acharkq@gmail.com}, 
\texttt{shuwen@stu.scu.edu.cn}, \\
\texttt{liang.wang@cripac.ia.ac.cn},
\small \texttt{luoyanchen@mail.ustc.edu.cn}, 
\texttt{wenqianglei@scu.edu.cn}, \\
\texttt{ybian@nus.edu.sg},
\small \texttt{fangjf1997@gmail.com}, 
\texttt{xiangwang1223@gmail.com} \\[2mm]
\small $^*$ Equal contribution. \quad $^\dagger$ Corresponding author.
}

\begin{document}

\maketitle

\begin{abstract} 
  Masked graph modeling (MGM) is a promising approach for molecular representation learning (MRL).
  \liang{However, extending the success of re-mask decoding from 2D to 3D MGM is non-trivial, primarily due to two conflicting challenges: avoiding 2D structure leakage to the decoder, while still providing sufficient 2D context for reconstructing re-masked atoms.}
  To address these challenges, we propose \textbf{3D-GSRD}: a \textbf{3D} Molecular \textbf{G}raph Auto-Encoder with \textbf{S}elective \textbf{R}e-mask \textbf{D}ecoding. The core innovation of 3D-GSRD lies in its  \textbf{S}elective \textbf{R}e-mask \textbf{D}ecoding (\textbf{SRD}), which re-masks only 3D-relevant information from encoder representations while preserving the 2D graph structures. 
  This SRD is synergistically integrated with  a \textbf{3D} \textbf{Re}lational-\textbf{Trans}former (\textbf{3D-ReTrans}) encoder alongside a structure-independent decoder. We analyze that SRD, combined with the structure-independent decoder, enhances the encoder's role in MRL. Extensive experiments show that 3D-GSRD achieves strong downstream performance, setting a new state-of-the-art on 7 out of 8 targets in the widely used MD17 molecular property prediction benchmark. The code is released at \url{https://github.com/WuChang0124/3D-GSRD}.
\end{abstract}

\input{sections/1_introduction_new}

\input{sections/2_related_work}

\input{sections/3_method}

\input{sections/4_analysis}

\input{sections/5_experiments}

\input{sections/6_conclusion}

\bibliographystyle{unsrtnat}
\bibliography{reference}


\newpage
\section*{NeurIPS Paper Checklist}

\begin{enumerate}

\item {\bf Claims}
    \item[] Question: Do the main claims made in the abstract and introduction accurately reflect the paper's contributions and scope?
    \item[] Answer: \answerYes{} 
    \item[] Justification: We have included the paper's contributions and scope in the abstract and introduction.
    \item[] Guidelines:
    \begin{itemize}
        \item The answer NA means that the abstract and introduction do not include the claims made in the paper.
        \item The abstract and/or introduction should clearly state the claims made, including the contributions made in the paper and important assumptions and limitations. A No or NA answer to this question will not be perceived well by the reviewers. 
        \item The claims made should match theoretical and experimental results, and reflect how much the results can be expected to generalize to other settings. 
        \item It is fine to include aspirational goals as motivation as long as it is clear that these goals are not attained by the paper. 
    \end{itemize}

\item {\bf Limitations}
    \item[] Question: Does the paper discuss the limitations of the work performed by the authors?
    \item[] Answer: \answerYes{}{} 
    \item[] Justification: The limitations are included in Section~\ref{sec:conclusion}.
    \item[] Guidelines:
    \begin{itemize}
        \item The answer NA means that the paper has no limitation while the answer No means that the paper has limitations, but those are not discussed in the paper. 
        \item The authors are encouraged to create a separate "Limitations" section in their paper.
        \item The paper should point out any strong assumptions and how robust the results are to violations of these assumptions (e.g., independence assumptions, noiseless settings, model well-specification, asymptotic approximations only holding locally). The authors should reflect on how these assumptions might be violated in practice and what the implications would be.
        \item The authors should reflect on the scope of the claims made, e.g., if the approach was only tested on a few datasets or with a few runs. In general, empirical results often depend on implicit assumptions, which should be articulated.
        \item The authors should reflect on the factors that influence the performance of the approach. For example, a facial recognition algorithm may perform poorly when image resolution is low or images are taken in low lighting. Or a speech-to-text system might not be used reliably to provide closed captions for online lectures because it fails to handle technical jargon.
        \item The authors should discuss the computational efficiency of the proposed algorithms and how they scale with dataset size.
        \item If applicable, the authors should discuss possible limitations of their approach to address problems of privacy and fairness.
        \item While the authors might fear that complete honesty about limitations might be used by reviewers as grounds for rejection, a worse outcome might be that reviewers discover limitations that aren't acknowledged in the paper. The authors should use their best judgment and recognize that individual actions in favor of transparency play an important role in developing norms that preserve the integrity of the community. Reviewers will be specifically instructed to not penalize honesty concerning limitations.
    \end{itemize}

\item {\bf Theory assumptions and proofs}
    \item[] Question: For each theoretical result, does the paper provide the full set of assumptions and a complete (and correct) proof?
    \item[] Answer: \answerNA{} 
    \item[] Justification: The paper does not include theoretical results. 
    \item[] Guidelines:
    \begin{itemize}
        \item The answer NA means that the paper does not include theoretical results. 
        \item All the theorems, formulas, and proofs in the paper should be numbered and cross-referenced.
        \item All assumptions should be clearly stated or referenced in the statement of any theorems.
        \item The proofs can either appear in the main paper or the supplemental material, but if they appear in the supplemental material, the authors are encouraged to provide a short proof sketch to provide intuition. 
        \item Inversely, any informal proof provided in the core of the paper should be complemented by formal proofs provided in appendix or supplemental material.
        \item Theorems and Lemmas that the proof relies upon should be properly referenced. 
    \end{itemize}

    \item {\bf Experimental result reproducibility}
    \item[] Question: Does the paper fully disclose all the information needed to reproduce the main experimental results of the paper to the extent that it affects the main claims and/or conclusions of the paper (regardless of whether the code and data are provided or not)?
    \item[] Answer: \answerYes{} 
    \item[] Justification: The information needed to reproduce the main experimental results is provided in Section~\ref{sec:experimental Setup} and Appendix~\ref{appendix:setup}.
    \item[] Guidelines:
    \begin{itemize}
        \item The answer NA means that the paper does not include experiments.
        \item If the paper includes experiments, a No answer to this question will not be perceived well by the reviewers: Making the paper reproducible is important, regardless of whether the code and data are provided or not.
        \item If the contribution is a dataset and/or model, the authors should describe the steps taken to make their results reproducible or verifiable. 
        \item Depending on the contribution, reproducibility can be accomplished in various ways. For example, if the contribution is a novel architecture, describing the architecture fully might suffice, or if the contribution is a specific model and empirical evaluation, it may be necessary to either make it possible for others to replicate the model with the same dataset, or provide access to the model. In general. releasing code and data is often one good way to accomplish this, but reproducibility can also be provided via detailed instructions for how to replicate the results, access to a hosted model (e.g., in the case of a large language model), releasing of a model checkpoint, or other means that are appropriate to the research performed.
        \item While NeurIPS does not require releasing code, the conference does require all submissions to provide some reasonable avenue for reproducibility, which may depend on the nature of the contribution. For example
        \begin{enumerate}
            \item If the contribution is primarily a new algorithm, the paper should make it clear how to reproduce that algorithm.
            \item If the contribution is primarily a new model architecture, the paper should describe the architecture clearly and fully.
            \item If the contribution is a new model (e.g., a large language model), then there should either be a way to access this model for reproducing the results or a way to reproduce the model (e.g., with an open-source dataset or instructions for how to construct the dataset).
            \item We recognize that reproducibility may be tricky in some cases, in which case authors are welcome to describe the particular way they provide for reproducibility. In the case of closed-source models, it may be that access to the model is limited in some way (e.g., to registered users), but it should be possible for other researchers to have some path to reproducing or verifying the results.
        \end{enumerate}
    \end{itemize}

\item {\bf Open access to data and code}
    \item[] Question: Does the paper provide open access to the data and code, with sufficient instructions to faithfully reproduce the main experimental results, as described in supplemental material?
    \item[] Answer: \answerYes{} 
    \item[] Justification: The code of the paper is released at \url{https://github.com/WuChang0124/3D-GSRD} and the data used are publicly available.
    \item[] Guidelines: 
    \begin{itemize}
        \item The answer NA means that paper does not include experiments requiring code.
        \item Please see the NeurIPS code and data submission guidelines (\url{https://nips.cc/public/guides/CodeSubmissionPolicy}) for more details.
        \item While we encourage the release of code and data, we understand that this might not be possible, so “No” is an acceptable answer. Papers cannot be rejected simply for not including code, unless this is central to the contribution (e.g., for a new open-source benchmark).
        \item The instructions should contain the exact command and environment needed to run to reproduce the results. See the NeurIPS code and data submission guidelines (\url{https://nips.cc/public/guides/CodeSubmissionPolicy}) for more details.
        \item The authors should provide instructions on data access and preparation, including how to access the raw data, preprocessed data, intermediate data, and generated data, etc.
        \item The authors should provide scripts to reproduce all experimental results for the new proposed method and baselines. If only a subset of experiments are reproducible, they should state which ones are omitted from the script and why.
        \item At submission time, to preserve anonymity, the authors should release anonymized versions (if applicable).
        \item Providing as much information as possible in supplemental material (appended to the paper) is recommended, but including URLs to data and code is permitted.
    \end{itemize}

\item {\bf Experimental setting/details}
    \item[] Question: Does the paper specify all the training and test details (e.g., data splits, hyperparameters, how they were chosen, type of optimizer, etc.) necessary to understand the results?
    \item[] Answer: \answerYes{} 
    \item[] Justification: The experimental settings are provided in Section~\ref{sec:experimental Setup} and Appendix~\ref{appendix:setup}.
    \item[] Guidelines:
    \begin{itemize}
        \item The answer NA means that the paper does not include experiments.
        \item The experimental setting should be presented in the core of the paper to a level of detail that is necessary to appreciate the results and make sense of them.
        \item The full details can be provided either with the code, in appendix, or as supplemental material.
    \end{itemize}

\item {\bf Experiment statistical significance}
    \item[] Question: Does the paper report error bars suitably and correctly defined or other appropriate information about the statistical significance of the experiments?
    \item[] Answer: \answerNo{} 
    \item[] Justification: We follow the experimental design of previous works, which did not involve repeated runs and did not report error bars. And due to limited computational resources, we are unable to afford the overhead of running multiple trials for each experiment.
    \item[] Guidelines:
    \begin{itemize}
        \item The answer NA means that the paper does not include experiments.
        \item The authors should answer "Yes" if the results are accompanied by error bars, confidence intervals, or statistical significance tests, at least for the experiments that support the main claims of the paper.
        \item The factors of variability that the error bars are capturing should be clearly stated (for example, train/test split, initialization, random drawing of some parameter, or overall run with given experimental conditions).
        \item The method for calculating the error bars should be explained (closed form formula, call to a library function, bootstrap, etc.)
        \item The assumptions made should be given (e.g., Normally distributed errors).
        \item It should be clear whether the error bar is the standard deviation or the standard error of the mean.
        \item It is OK to report 1-sigma error bars, but one should state it. The authors should preferably report a 2-sigma error bar than state that they have a 96\% CI, if the hypothesis of Normality of errors is not verified.
        \item For asymmetric distributions, the authors should be careful not to show in tables or figures symmetric error bars that would yield results that are out of range (e.g. negative error rates).
        \item If error bars are reported in tables or plots, The authors should explain in the text how they were calculated and reference the corresponding figures or tables in the text.
    \end{itemize}

\item {\bf Experiments compute resources}
    \item[] Question: For each experiment, does the paper provide sufficient information on the computer resources (type of compute workers, memory, time of execution) needed to reproduce the experiments?
    \item[] Answer: \answerYes{} 
    \item[] Justification: The information about compute resources are provided in Appendix~\ref{appendix:setup}.
    \item[] Guidelines:
    \begin{itemize}
        \item The answer NA means that the paper does not include experiments.
        \item The paper should indicate the type of compute workers CPU or GPU, internal cluster, or cloud provider, including relevant memory and storage.
        \item The paper should provide the amount of compute required for each of the individual experimental runs as well as estimate the total compute. 
        \item The paper should disclose whether the full research project required more compute than the experiments reported in the paper (e.g., preliminary or failed experiments that didn't make it into the paper). 
    \end{itemize}
    
\item {\bf Code of ethics}
    \item[] Question: Does the research conducted in the paper conform, in every respect, with the NeurIPS Code of Ethics \url{https://neurips.cc/public/EthicsGuidelines}?
    \item[] Answer: \answerYes{} 
    \item[] Justification: The research in this paper conform with the NeurIPS Code of Ethics.
    \item[] Guidelines:
    \begin{itemize}
        \item The answer NA means that the authors have not reviewed the NeurIPS Code of Ethics.
        \item If the authors answer No, they should explain the special circumstances that require a deviation from the Code of Ethics.
        \item The authors should make sure to preserve anonymity (e.g., if there is a special consideration due to laws or regulations in their jurisdiction).
    \end{itemize}

\item {\bf Broader impacts}
    \item[] Question: Does the paper discuss both potential positive societal impacts and negative societal impacts of the work performed?
    \item[] Answer: \answerYes{} 
    \item[] Justification: We discuss the societal impacts of the work performed in Appendix~\ref{appendix:Broader Impacts}.
    \item[] Guidelines:
    \begin{itemize}
        \item The answer NA means that there is no societal impact of the work performed.
        \item If the authors answer NA or No, they should explain why their work has no societal impact or why the paper does not address societal impact.
        \item Examples of negative societal impacts include potential malicious or unintended uses (e.g., disinformation, generating fake profiles, surveillance), fairness considerations (e.g., deployment of technologies that could make decisions that unfairly impact specific groups), privacy considerations, and security considerations.
        \item The conference expects that many papers will be foundational research and not tied to particular applications, let alone deployments. However, if there is a direct path to any negative applications, the authors should point it out. For example, it is legitimate to point out that an improvement in the quality of generative models could be used to generate deepfakes for disinformation. On the other hand, it is not needed to point out that a generic algorithm for optimizing neural networks could enable people to train models that generate Deepfakes faster.
        \item The authors should consider possible harms that could arise when the technology is being used as intended and functioning correctly, harms that could arise when the technology is being used as intended but gives incorrect results, and harms following from (intentional or unintentional) misuse of the technology.
        \item If there are negative societal impacts, the authors could also discuss possible mitigation strategies (e.g., gated release of models, providing defenses in addition to attacks, mechanisms for monitoring misuse, mechanisms to monitor how a system learns from feedback over time, improving the efficiency and accessibility of ML).
    \end{itemize}
    
\item {\bf Safeguards}
    \item[] Question: Does the paper describe safeguards that have been put in place for responsible release of data or models that have a high risk for misuse (e.g., pretrained language models, image generators, or scraped datasets)?
    \item[] Answer: \answerNo{} 
    \item[] Justification: The paper poses no such risks.
    \item[] Guidelines:
    \begin{itemize}
        \item The answer NA means that the paper poses no such risks.
        \item Released models that have a high risk for misuse or dual-use should be released with necessary safeguards to allow for controlled use of the model, for example by requiring that users adhere to usage guidelines or restrictions to access the model or implementing safety filters. 
        \item Datasets that have been scraped from the Internet could pose safety risks. The authors should describe how they avoided releasing unsafe images.
        \item We recognize that providing effective safeguards is challenging, and many papers do not require this, but we encourage authors to take this into account and make a best faith effort.
    \end{itemize}

\item {\bf Licenses for existing assets}
    \item[] Question: Are the creators or original owners of assets (e.g., code, data, models), used in the paper, properly credited and are the license and terms of use explicitly mentioned and properly respected?
    \item[] Answer: \answerYes{} 
    \item[] Justification: The PCQM4Mv2 dataset are used by \url{https://ogb.stanford.edu/docs/lsc/pcqm4mv2/} under CC BY 4.0 License. The QM9 dataset are used via \url{https://deepchemdata.s3-us-west-1.amazonaws.com/} under the MIT License.  
    \item[] Guidelines:
    \begin{itemize}
        \item The answer NA means that the paper does not use existing assets.
        \item The authors should cite the original paper that produced the code package or dataset.
        \item The authors should state which version of the asset is used and, if possible, include a URL.
        \item The name of the license (e.g., CC-BY 4.0) should be included for each asset.
        \item For scraped data from a particular source (e.g., website), the copyright and terms of service of that source should be provided.
        \item If assets are released, the license, copyright information, and terms of use in the package should be provided. For popular datasets, \url{paperswithcode.com/datasets} has curated licenses for some datasets. Their licensing guide can help determine the license of a dataset.
        \item For existing datasets that are re-packaged, both the original license and the license of the derived asset (if it has changed) should be provided.
        \item If this information is not available online, the authors are encouraged to reach out to the asset's creators.
    \end{itemize}

\item {\bf New assets}
    \item[] Question: Are new assets introduced in the paper well documented and is the documentation provided alongside the assets?
    \item[] Answer: \answerNA{} 
    \item[] Justification: The paper does not release new assets.
    \item[] Guidelines:
    \begin{itemize}
        \item The answer NA means that the paper does not release new assets.
        \item Researchers should communicate the details of the dataset/code/model as part of their submissions via structured templates. This includes details about training, license, limitations, etc. 
        \item The paper should discuss whether and how consent was obtained from people whose asset is used.
        \item At submission time, remember to anonymize your assets (if applicable). You can either create an anonymized URL or include an anonymized zip file.
    \end{itemize}

\item {\bf Crowdsourcing and research with human subjects}
    \item[] Question: For crowdsourcing experiments and research with human subjects, does the paper include the full text of instructions given to participants and screenshots, if applicable, as well as details about compensation (if any)? 
    \item[] Answer: \answerNA{} 
    \item[] Justification: The paper does not involve crowdsourcing nor research with human subjects.
    \item[] Guidelines:
    \begin{itemize}
        \item The answer NA means that the paper does not involve crowdsourcing nor research with human subjects.
        \item Including this information in the supplemental material is fine, but if the main contribution of the paper involves human subjects, then as much detail as possible should be included in the main paper. 
        \item According to the NeurIPS Code of Ethics, workers involved in data collection, curation, or other labor should be paid at least the minimum wage in the country of the data collector. 
    \end{itemize}

\item {\bf Institutional review board (IRB) approvals or equivalent for research with human subjects}
    \item[] Question: Does the paper describe potential risks incurred by study participants, whether such risks were disclosed to the subjects, and whether Institutional Review Board (IRB) approvals (or an equivalent approval/review based on the requirements of your country or institution) were obtained?
    \item[] Answer: \answerNA{} 
    \item[] Justification: The paper does not involve crowdsourcing nor research with human subjects.
    \item[] Guidelines:
    \begin{itemize}
        \item The answer NA means that the paper does not involve crowdsourcing nor research with human subjects.
        \item Depending on the country in which research is conducted, IRB approval (or equivalent) may be required for any human subjects research. If you obtained IRB approval, you should clearly state this in the paper. 
        \item We recognize that the procedures for this may vary significantly between institutions and locations, and we expect authors to adhere to the NeurIPS Code of Ethics and the guidelines for their institution. 
        \item For initial submissions, do not include any information that would break anonymity (if applicable), such as the institution conducting the review.
    \end{itemize}

\item {\bf Declaration of LLM usage}
    \item[] Question: Does the paper describe the usage of LLMs if it is an important, original, or non-standard component of the core methods in this research? Note that if the LLM is used only for writing, editing, or formatting purposes and does not impact the core methodology, scientific rigorousness, or originality of the research, declaration is not required.
    \item[] Answer: \answerNA{} 
    \item[] Justification: The core method development in this research does not involve LLMs as any important, original, or non-standard components.
    \item[] Guidelines:
    \begin{itemize}
        \item The answer NA means that the core method development in this research does not involve LLMs as any important, original, or non-standard components.
        \item Please refer to our LLM policy (\url{https://neurips.cc/Conferences/2025/LLM}) for what should or should not be described.
    \end{itemize}

\end{enumerate}

\newpage
\appendix
\input{sections/8_appendix}

\clearpage

\end{document}

%% file: sections/1_introduction_new.tex
\section{Introduction}
\label{Introduction}


Molecular representation learning (MRL) \cite{fang2022geometry, zhang2024subgdiff, ji2022relmole} is fundamental to a wide range of downstream tasks, including 
de novo drug design \cite{wang2022deep}, molecular dynamics simulation \cite{bai2022application}, and molecular property prediction \cite{li2022geomgcl, stark20223d}. Given the abundance of unlabeled molecular data in this field, self-supervised pretraining has emerged as a key strategy for learning effective molecular representations.
Previous works have primarily focused on 1D molecular strings \cite{feng2023unimap,krenn2020self} and 2D molecular graphs \cite{xia2023mole, sun2021mocl, wang2022improving}, achieving promising results. However, they often neglect critical 3D structural information, which is crucial for capturing molecular properties such as the highest occupied molecular orbital, molecular dynamics, and energy functions \cite{qm9}. This limitation has led to a growing interest in incorporating 3D molecular coordinates into pretraining frameworks.



\begin{figure}[t]
\small
\centering
\begin{minipage}{0.48\textwidth}
    \centering
    \includegraphics[width=\textwidth]{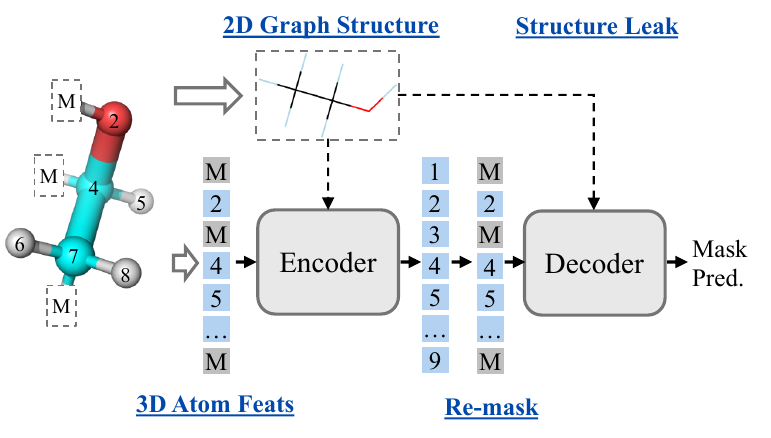}
    \caption{Illustration of 3D MGM with re-mask decoding and 2D structure leakage.}
    \vspace{-10pt}
    \label{fig:3d_mgm}
\end{minipage}\hfill
\begin{minipage}{0.45\textwidth}
    \centering
    \includegraphics[width=\textwidth]{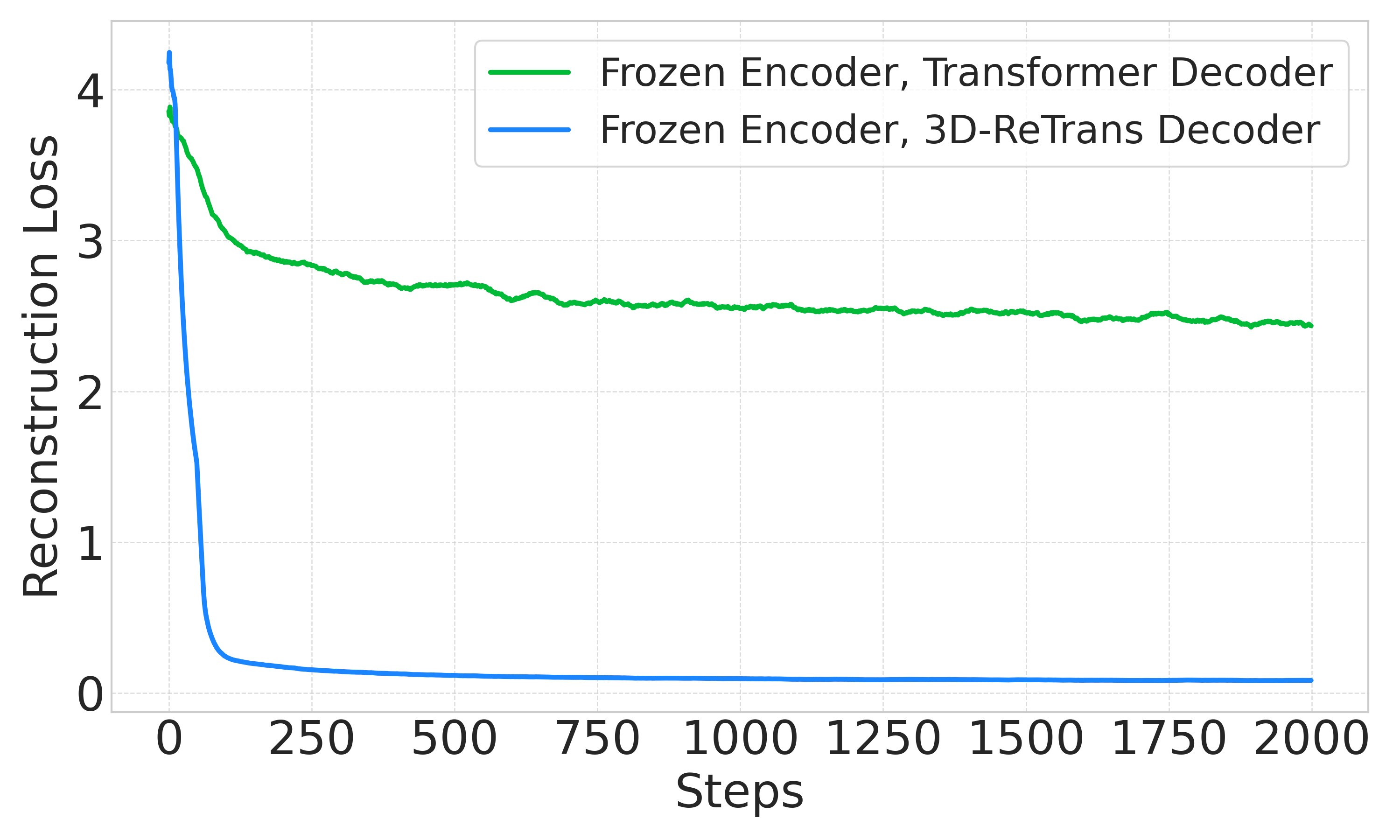}
    \vspace{-15pt}
    \caption{Reconstruction loss across two pretraining settings. We compare settings using frozen encoders (3D-ReTrans) and structure-independent (Transformer) versus structure-dependent (3D-ReTrans) decoders.}
    \vspace{-10pt}
    \label{fig:intro-analysis1}
\end{minipage}\hfill
\end{figure}

\liang{Masked graph modeling (MGM) has emerged as a leading paradigm for 3D molecular pretraining, aiming to learn data distributions by reconstructing randomly masked graph features \cite{unimol,graphmae}.}
As illustrated in \cref{fig:3d_mgm}, its 3D variant typically consists of three key components: (1) 3D graph masking, which perturbs the original 3D molecular graph by randomly masking features such as 3D coordinates, atom types, and chemical bonds \cite{unimol,unimol2}; (2) a 3D graph encoder, which processes the masked graph to generate molecular representations; and (3) a 3D graph decoder, which reconstructs the masked features from the encoded representations. After pretraining via reconstruction, the 3D graph encoder is finetuned on downstream tasks to enhance performance.




However, a long-standing challenge in MGM, mirroring similar issues in masked image modeling \cite{mae}, is the misalignment between the reconstruction objective and representation learning \cite{simsgt}. Specifically, minimizing the reconstruction error often leads models to focus on low-level graph features, such as atom types and 3D coordinates, rather than learning high-level graph semantics required for property prediction \cite{molae}. To mitigate this, re-mask decoding is introduced for MGM pretraining on 2D molecular graphs \cite{graphmae,graphmae2,simsgt}. This method re-masks the encoder's representation of previously masked atoms before feeding them to the decoder (\cref{fig:3d_mgm}). In this way, the encoder is prevented from reconstructing the masked atoms directly and is encouraged to focus on generating high-quality representations of the unmasked graph regions, which the decoder then uses for reconstruction. This approach shifts the encoder's focus from graph reconstruction to MRL and yields substantial downstream improvements for 2D MGM.

To adapt re-mask decoding for 3D MGM, we identify two seemingly contradictory challenges:
\begin{itemize}[leftmargin=*]
\item \liang{\textbf{Leaking 2D structure to decoder weakens encoder's MRL capability.}}
The decoder should rely solely on the encoder representations to reconstruct masked features. Exposing the decoder directly to 2D molecular structures (\eg chemical bond connections) diminishes the encoder's role in MRL, as the decoder can recover masked features using the provided 2D molecular structures, even with poor encoder representations. For example, \cref{fig:intro-analysis1} shows that a frozen randomly-initialized encoder with a trainable structure-dependent decoder can achieve relatively low reconstruction loss when predicting masked atomic coordinates. 
Consequently, the encoder focuses less on capturing structures, leading to suboptimal MRL performance. However, existing re-mask decoding methods \cite{graphmae,simsgt,graphmae2} typically use structure-dependent decoders like graph neural networks \cite{GIN}, which exacerbate this issue.
\item \liang{\textbf{Structure-independent decoding can prevent structure leakage, but hinders reconstruction of re-masked atoms.}}
A naive solution to prevent structure leakage is to use a structure-independent decoder, which consumes no 2D structure input. However, this approach fails to account for the relative positions and contextual relationships of re-masked atoms within the 2D graph, making it challenging to distinguish between re-masked atoms during reconstruction. To address this, we propose leveraging the 3D graph encoder to generate 2D structural contexts for re-masked atoms. This ensures that the encoder is effectively trained for structural representation 
\yuan{while preventing structure leakage beyond the encoder's representations.}

\end{itemize}


To address the challenges above, we introduce \textbf{3D} Molecular \textbf{G}raph Auto-Encoder with \textbf{S}elective \textbf{R}e-mask \textbf{D}ecoding (\textbf{3D-GSRD}), a 3D MGM framework with three key elements: (1) the \textbf{S}elective \textbf{R}e-mask \textbf{D}ecoding (\textbf{SRD}) that re-masks only 3D-relevant information from the encoder representations while preserving its 2D structural context; (2) a structure-independent decoder that derives all structural information exclusively from the encoder; and (3) \yuan{\textbf{3D} \textbf{Re}lational-\textbf{Trans}former (termed as \textbf{3D-ReTrans}) as 3D graph encoder, effectively integrating 3D molecular features (\eg atomic coordinates) and 2D features (\eg bonding connections) for MRL.}

Specifically, SRD ensures the preservation of 2D graph structures during re-masking by reintroducing 2D information through a \textbf{2D} graph \textbf{P}osition \textbf{E}ncoder (\textbf{2D-PE}).
Crucially, this 2D-PE is trained via distillation from the 3D graph encoder's representations, ensuring its information is fully contained within the 3D encoder, as demonstrated in Section~\ref{sec:analysis}. The distillation process also allows the 2D-PE's structural encoding capability to improve alongside the 3D encoder's advancements during pretraining.
To complement SRD, we employ a structure-independent Transformer \cite{transformer} decoder that derives 2D graph structure exclusively from \yuan{the 3D encoder and the 2D-PE}. 
Together with SRD, our decoder provides the re-masked atoms with rich 2D graph contexts that are distilled from the 3D graph encoder, while preventing 2D structure leakage.

Encoding 3D molecules is challenging due to their multi-modal nature (\eg discrete atom types vs. continuous coordinates) and multi-granular structure (\eg atom-wise vs. pair-wise features). Prior works like PaiNN \cite{painn} and TorchMD-NET \cite{torchmd} address this by using equivariant architectures that separately process scalar features (\eg atom types, distances) and vector features (\eg directional geometry). Building on these insights, we introduce \textbf{3D-ReTrans} as our 3D graph encoder, extending the Relational-Transformer \cite{RelaTrans} to incorporate both scalar and vector features while maintaining its scalability and flexibility to process both atom-wise and pair-wise features. Specifically, we introduce a tailored attention mechanism that incorporates pairwise distances and interactions directly into attention weights, along with a 3D Update Layer that jointly updates scalar and vector features. This design yields strong performance on MRL and serves as a robust backbone for 3D MGM pretraining.


Finally, we include in-depth analysis to showcase the inner mechanism of SRD and our structure-independent decoder, \yuan{demonstrating our key claims of shifting the encoder's focus to MRL while preventing structure leakage in the decoder.} Based on these revealed advantages in MGM pretraining, 3D-GSRD demonstrates superior performance when being fine-tuned on downstream datasets, achieving new state-of-the-art on 7 out of 8 molecules for MD17 \cite{md17}.

%% file: sections/2_related_work.tex
\section{Related Work}
\label{Related Work}
Molecular pretraining has emerged as a fundamental approach for molecular representation learning \cite{rong2020self,li2020learn,graphtransformersurvey}, critical for various downstream tasks, such as molecule property prediction.

\textbf{3D Molecular Denoising and Masked Graph Modeling.} 
Recent advances in 3D molecular pretraining have focused on 3D structure learning through coordinate denoising and masking.
For example, \cite{feng2023fractional,ni2023sliced,zaidi2022pre} introduce noise to atomic coordinates and then reconstruct them. SubGDiff \cite{zhang2024subgdiff} adds distinct Gaussian noise to different substructures of 3D molecular conformation and performs denoising via a diffusion process. 
\reb{MolSpectra \cite{wang2025molspectra} uses the energy spectra to enhance 3D molecular representation learning during denoising.} 
As for masking, Uni-Mol \cite{unimol} and Uni-Mol2 \cite{unimol2} employ masked coordinates prediction as one of the self-supervised tasks, while other works like \cite{fang2022geometry} focus on masking and predicting bond lengths and angles.

\textbf{Other 3D Molecular Pretraining Methods.} 
EPT \cite{jiao2024equivariant} proposes a multi-domain 3D pretraining approach by combining atom-level features for small molecules and residue-level features in proteins. 3D PGT \cite{wang2023automated} designs three generative pretraining tasks, including predicting bond length, bond angle, and dihedral angle, and introduces an adaptive fusion strategy for these tasks, using total energy as a surrogate metric to optimize their combination weight. \reb{GraphMVP \cite{liu2021pre} and 3D Infomax \cite{stark20223d} use contrastive learning to transfer knowledge from the 3D encoder into the 2D graph encoder.}

\textbf{Graph Position Encoding and Structure Encoding.}  
Position Encoding (PE) encodes the spatial position of a given node within a graph \cite{GraphGPS}. Some methods \cite{kreuzer2021rethinking,dwivedi2023benchmarking} use adjacency, Laplacian, or distance matrices to represent PE. Other approaches like \cite{li2020distance,mialon2021graphit,beaini2021directional} leverage shortest paths, heat kernels, or Green's function to compute pair-wise distance, capturing the distance and directional relationships between nodes. Currently, in MOL-AE \cite{molae}, SMILES strings are used to provide PE to the decoder as an identifier. 
Structure Encoding (SE) encodes the structural information of graphs and subgraphs. Common methods include node degree \cite{ying2021transformers}, Laplacian matrices \cite{kreuzer2021rethinking}, and Boolean indicators that specify whether two nodes belong to the same substructure \cite{bodnar2021weisfeiler}. Unlike these methods, we propose using a 2D graph position encoder distilled from a 3D graph encoder to produce SE, which provides rich and effective context information.

\textbf{2D Molecular Graph Pretraining.} Previous methods primarily focus on leveraging 2D molecular graphs to learn molecular representation. A popular technique is masked graph modeling \cite{graphmae,feng2023unimap,xia2023mole}, typically comprising three key components \cite{simsgt}: graph tokenizer \cite{sun2021mocl,ji2022relmole}, graph masking \cite{you2020does,hu2019strategies}, and graph autoencoder \cite{li2022maskgae,wang2017mgae,zhang2022graph}. Another prominent line of work is contrastive learning \cite{zhu2023dual,wang2022improving,fang2022molecular}, which aims to pull positive pairs and push negative pairs apart in the representation space. Notably, methods such as GeomGCL \cite{li2022geomgcl}, GraphMVP \cite{liu2021pre}, and 3D Infomax \cite{stark20223d} incorporate 3D molecular conformations as auxiliary information to enhance 2D graph representations via contrastive objectives. While effective for 2D molecular pretraining, these methods overlook 3D features, which are crucial for molecular representation learning. Moreover, directly extending these methods to 3D molecular pretraining is non-trivial due to the increased complexity and spatial nature of 3D molecular data.

\reb{Our method is similar to GraphMVP \cite{liu2021pre} in leveraging both 2D and 3D molecular graphs but differs in objective and design. While GraphMVP transfers 3D information into a 2D encoder to enhance 2D graph representations, our method distills 3D representation into 2D-PE, ensuring the 2D-PE's embedding is fully contained within the 3D encoder to avoid structure leakage in decoding. Additionally, GraphMVP aligns the 2D and 3D views of the same molecule and contrasts views of different molecules using contrastive losses. Our method instead uses a cosine similarity loss that encourages 2D-PE to generate structural encodings closely aligned with the 3D encoder. This offers a simpler and more efficient framework for 2D structure-informed decoding without structure leakage.}

%% file: sections/3_method.tex
\section{Preliminary: 3D Masked Graph Modeling}
\textbf{Notations.} A 3D molecular graph is represented as $G=(\mathbf{x},\mathbf{a},\mathbf{e})$, where $\mathbf{x}\in \mathbb{R}^{N\times 3}$ denotes the 3D atomic coordinates, 
$\mathbf{a}\in \mathbb{R}^{N\times *}$ represents the atom types, and $\mathbf{e}\in \mathbb{R}^{N\times N\times *}$ captures the atomic pair features, such as inter-atomic distances and bonds. $N$ is the number of atoms.

\textbf{Graph Masking.} Given a molecular graph $G$, the 3D coordinates $\{\mathbf{x}_i \in \mathbb{R}^3 | i\in \mathcal{V}_m\}$ of a randomly selected subset of atoms $\mathcal{V}_{m}$ are masked. For each masked atom $i\in \mathcal{V}_{m}$, its original coordinates $\mathbf{x}_i$ is replaced by a learnable special token $\mathbf{m}_x \in \mathbb{R}^{3}$. 
The coordinate matrix $\mathbf{x}$ after masking is denoted as $\tilde{\mathbf{x}}$, and the masked graph is denoted as $\tilde{G}=(\tilde{\mathbf{x}}, \mathbf{a},\mathbf{e})$. 
\wuc{Some prior works \cite{molae} instead remove all information corresponding to the masked atoms, including their 3D coordinates, atom types, and pairwise features, which results in a masked graph $\tilde{G} = (\tilde{\mathbf{x}}, \tilde{\mathbf{a}}, \tilde{\mathbf{e}})$. In this work, we adopt this latter masking strategy, fully excluding the masked atoms from the input graph.}

\textbf{3D Graph Auto-Encoder and 2D Structure Leakage.} The 3D graph auto-encoder comprises a graph encoder $\phi_e(\cdot)$ and a graph decoder $\phi_d(\cdot)$. The encoder processes the masked graph $\tilde{G}$ to produce graph representations $\mathbf{h}=\phi_e(\tilde{G})\in \mathbb{R}^{N\times *}$. 
The decoder then predicts the masked coordinates $\{\hat{\mathbf{x}}_i|i\in \mathcal{V}_m\}$ using $\phi_d(\mathbf{h})$, and optionally incorporating the pair features $\phi_d(\mathbf{h},\mathbf{e})$. However, using pair features $\mathbf{e}$ introduces \textbf{2D structure leakage}, as the decoder relies on additional information beyond the encoder's representation $\mathbf{h}$. \yuan{This weakens the encoder's role in MRL, because the decoder can leverage pair features to compensate for any deficiencies in the encoder's representations.} Despite this drawback, such leakage is common in previous MGM works \cite{graphmae,graphmae2,simsgt} that utilize Graph Neural Networks \cite{GIN} as decoders. In contrast, methods that avoid 2D structure leakage \cite{unimol,unimol2} mostly use weak decoders, such as MLPs, which can lead to suboptimal MGM pretraining.

\begin{figure*}[t]
    \small
    \centering
    \includegraphics[width=1.0\textwidth]{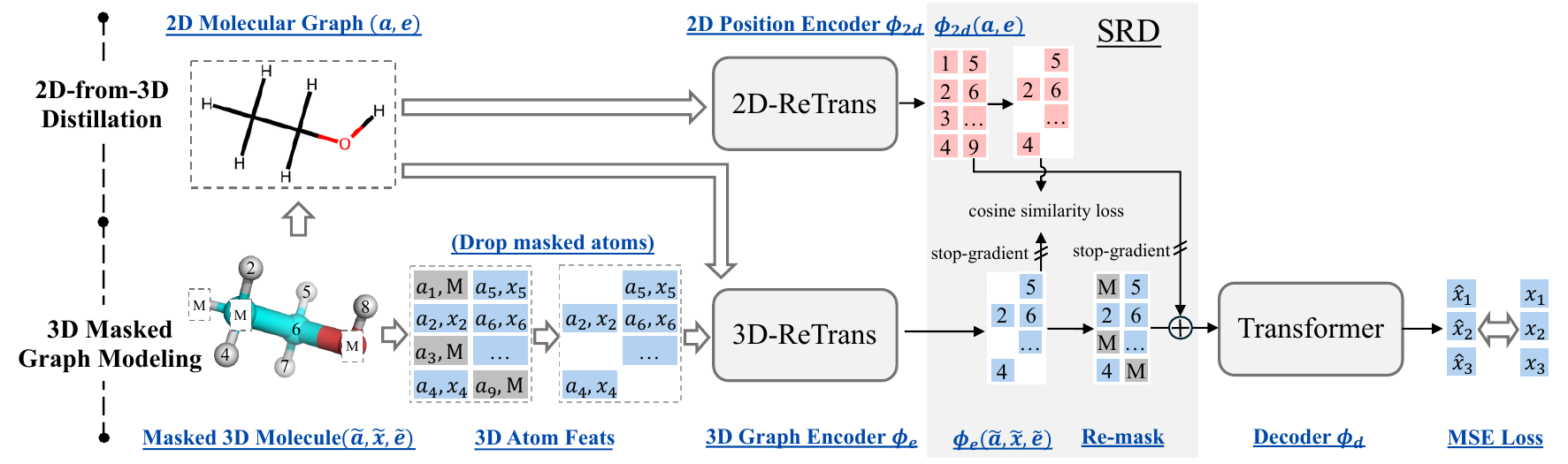}
    \vspace{-5mm}
    \caption{Overview of 3D-GSRD. It contains three key elements: (1) a 3D-ReTrans encoder; (2) the SRD that re-masks only 3D-relevant information from the encoder representations while preserving its 2D structure information via 2D-from-3D distillation; (3) a structure-independent decoder.}
     \vspace{-5mm}
    \label{fig:masked_graph_modeling}
\end{figure*}

\textbf{Re-mask Decoding.} Before passing $\mathbf{h}$ into the decoder, re-mask decoding replaces the representations of the previously masked atoms $\mathcal{V}_{m}$ with a learnable token $\mathbf{m}_h$, preventing the encoder from directly predicting the masked coordinates. This encourages the encoder to focus on learning meaningful representations for the unmasked graph regions. The re-masked representation $\tilde{\mathbf{h}}$ is defined as:
\begin{equation}
\tilde{\mathbf{h}}_i=\text{re-mask}(\mathbf{h}_i)=\begin{cases}
    \mathbf{m}_h, \hfill & \forall i\in \mathcal{V}_m, \\
    \mathbf{h}_i, \hfill & \text{otherwise}.
\end{cases}
\end{equation}

\textbf{MGM Loss.} The pretraining objective minimizes the mean squared error between the ground truth coordinates  $\{\mathbf{x}_i | i \in \mathcal{V}_m\}$  of the masked atoms and the decoder's predicted coordinates  $\{\hat{\mathbf{x}}_i | i \in \mathcal{V}_m\}$:
\begin{equation}
\mathcal{L}_{\text{MGM}} = \sum_{i\in \mathcal{V}_m} \|\hat{\mathbf{x}}_i - \mathbf{x}_i\|^2.
\end{equation}

\section{Methodology: 3D-GSRD}
In this section, we present our method \textbf{3D} Molecular \textbf{G}raph Auto-Encoder with \textbf{S}elective \textbf{R}e-mask \textbf{D}ecoding (\textbf{3D-GSRD}) (Figure~\ref{fig:masked_graph_modeling}). Below, we start by elaborating on the Selective Re-mask Decoding and the pretraining objective of 3D-GSRD. We then describe the encoder of 3D-ReTrans.

\begin{figure}[t]
    \centering
    \small
    \includegraphics[width=1.0\linewidth]{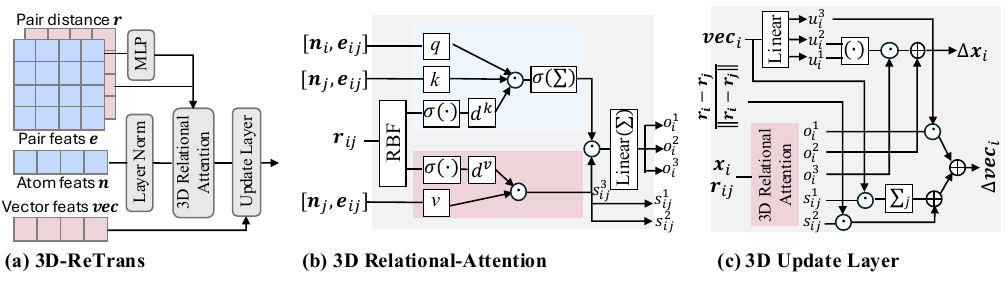}
    \vspace{-5mm}
    \caption{Illustration of 3D-ReTrans. \textbf{(a)} 3D-ReTrans is constructed by stacking multiple 3D Relational-Attention and 3D Update Layers. \textbf{(b)} 3D Relational-Attention that processes both atom-wise and pair-wise features. \textbf{(c)} 3D Update Layer that includes a residual connection.}
    \label{fig:3dreltrans}
    \vspace{-5mm}
\end{figure}

\subsection{SRD: Selective Re-mask Decoding}
Here we introduce SRD to improve 3D MGM. Re-mask decoding is proposed to address the mismatch between the reconstruction objective of 2D MGM and MRL \cite{graphmae,simsgt}. Our SRD extends this approach to 3D MGM while overcoming issues such as 2D structure leakage and providing 2D contexts to re-masked atoms. 

\reb{\textbf{Re-mask Decoding with 2D Graph Position Encoder}}. As Figure~\ref{fig:masked_graph_modeling} shows, given the encoder representation $\mathbf{h}=\phi_{e}(\tilde{\mathbf{x}},\tilde{\mathbf{a}},\tilde{\mathbf{e}})$ of $\tilde{G}$, SRD can be defined as:
\begin{equation}
\text{SRD}(\mathbf{h}, \tilde{G})=\text{re-mask}(\mathbf{h}) + \text{stop-grad}(\phi_{2d}(\mathbf{a},\mathbf{e})),
\end{equation}
where $\text{SRD}(\mathbf{h}, \tilde{G})$ is directly fed into the decoder for masked prediction; $\text{re-mask}(\cdot)$ is the standard re-mask; and $\text{stop-grad}(\cdot)$ stops the gradient flow to the 2D graph position encoder $\phi_{2d}$, which generates $\tilde{G}$'s 2D representation $\phi_{2d}(\mathbf{a},\mathbf{e})\in \mathbb{R}^{N\times *}$. 

\textbf{Building a 2D Graph Position Encoder without Structure Leakage via 2D-from-3D Distillation.} The 2D graph position encoder $\phi_{2d}(\mathbf{a},\mathbf{e})$ is the key component of SRD. \yuan{For unmasked atoms, $\phi_{2d}$ conveys the same information as  the 3D graph encoder $\phi_{e}$, preventing any information leakage beyond what $\phi_{e}$ has captured. For re-masked atoms, $\phi_{2d}$ offers the necessary 2D contexts that would have been available from $\phi_{e}$ without re-masking,} enabling the decoder to distinguish the relative positions of re-masked atoms. To prevent information leakage, $\phi_{2d}$ is trained exclusively through knowledge distillation from the 3D encoder, without any gradient updates from the MGM loss. This is enforced by the $\text{stop-grad}(\cdot)$, which blocks the gradient flow from the MGM loss into $\phi_{2d}$. 
Further, the knowledge distillation loss for $\phi_{2d}$ can be written as:
\begin{equation}
\mathcal{L}_{\text{distill}} = -\sum_{i\notin \mathcal{V}_m} \text{cos}(\phi_{2d}(\mathbf{a}, \mathbf{e})_i, \text{stop-grad}(\phi_{e}(\mathbf{x},\mathbf{a},  \mathbf{e})_i)),
\end{equation}
where $\text{cos}(\cdot,\cdot)$ denotes cosine similarity. The loss applies only to unmasked atoms for the consistency of $\phi_{\text{2d}}$'s training objective. We employ $\text{stop-grad}(\cdot)$ to prevent updating the 3D graph encoder $\phi_e$, allowing it to focus on MRL.

While the 2D graph position encoder can be any graph encoder, we implement it as a 2D-ReTrans, a simplified version of the 3D-ReTrans that excludes 3D coordinates and distance inputs. Our experiments demonstrate the effectiveness of this design.

\subsection{Pretraining 3D-GSRD}

\textbf{3D Graph Auto-Encoder.} We employ the 3D-ReTrans as encoder and employ SRD with a structure-independent decoder of transformer \cite{transformer}. In this way, we avoid structure leakage to the decoder beyond the encoder's information and provide strong 2D structural contexts for the re-masked atoms.

\textbf{Pretraining Loss.} We combine the MGM loss and 2D-from-3D distillation loss for pretraining:
\vspace{-3pt}
\begin{equation}
\mathcal{L}_{\text{pretrain}} = \mathcal{L}_{\text{MGM}} + \mathcal{L}_{\text{distill}}.
\end{equation}
\vspace{-18pt}

\subsection{3D Relational-Transformer}
Encoding 3D molecular graphs $G = (\mathbf{x},\mathbf{a},\mathbf{e})$ presents significant challenges of processing 3D coordinates $\mathbf{x}$ while preserving 3D equivariance and integrating the pairwise features $\mathbf{e}$, whose shape $(N,N,*)$ differs from the atomic features $(N,*)$. 
Prior works have primarily focused on ensuring 3D equivariance while processing the 3D coordinates $\mathbf{x}$ \cite{torchmd,Transformer-M}, but have paid less attention to effectively incorporating pair features $\mathbf{e}$. Most existing methods \cite{unimol,Transformer-M,AF3} incorporate pairwise representations of scalar values in self-attention layers \cite{transformer}, limiting their ability to capture the high-dimensional nature of inter-atomic interactions. While TorchMD-NET \cite{torchmd} models distances as high-dimensional pairwise features, extending it to include chemical bonds remains challenging.

To address the challenges, we propose \textbf{3D-ReTrans} as our encoder, leveraging the Relational-Transformer's \cite{RelaTrans} scalability and flexibility to incorporate pair features, while enabling it to process 3D coordinates. 
Draw inspiration from prior works \cite{torchmd,painn,gvp}, a core design is to explicitly separate and jointly process two types of features: (1) scalar features, which encode scalar information like atom types and distances; (2) vector features, which capture directional geometric information. Based on this, our key enhancements focus on improving its attention mechanism and incorporating a 3D update layer. More details about 3D-ReTrans are provided in Appendix~\ref{appendix:method}.

\textbf{3D Relational-Attention.} Each atom is represented by concatenating its types and coordinates: $\mathbf{n}_i = [\mathbf{a}_i; \mathbf{x}_i]$. The interaction between atoms $i$ and $j$ is captured by the pair feature $\mathbf{e}_{ij} \in \mathbb{R}^d$ and their Euclidean distance $r_{ij}$. 3D Relational-Attention is defined as:
\vspace{-3mm}

\begin{tabular}{p{7cm}p{6.2cm}}
\begin{equation}
\mathbf{q}_{ij} = [\mathbf{n}_{i}, \mathbf{e}_{ij}]\mathbf{W}^q, 
\end{equation}
&
\begin{equation}
[\mathbf{k}_{ij}; \mathbf{v}_{ij}] = [\mathbf{n}_{j}; \mathbf{e}_{ij}][\mathbf{W}^k;\mathbf{W}^v],
\end{equation} \\
\vspace{-10mm}
\begin{equation}
[\mathbf{d}_{ij}^k; \mathbf{d}_{ij}^v] = \text{SiLU}\left([\mathbf{W}^{dk};\mathbf{W}^{dv}]e^{\text{RBF}}(r_{ij})\right),
\end{equation}
&
\vspace{-8mm}
\begin{equation}
[\mathbf{s}_{ij}^1, \mathbf{s}_{ij}^2, \mathbf{s}_{ij}^3] = \mathbf{v}_{ij} \odot \mathbf{d}_{ij}^v,
\end{equation} \\
\vspace{-9mm}
\begin{equation}
\boldsymbol{\alpha}_{ij} = \text{SiLU}_j(\frac{\mathbf{q}_{ij} \cdot (\mathbf{k}_{ij} \odot \mathbf{d}_{ij}^k)}{\sqrt{d}}),
\end{equation}
&
\vspace{-12mm}
\begin{equation}
[\mathbf{o}_{i}^1, \mathbf{o}_{i}^2, \mathbf{o}_{i}^3] = \mathbf{W}^f (\sum_{j=1}^N \boldsymbol{\alpha}_{ij} \mathbf{s}_{ij}^3),
\end{equation}
\vspace{-12mm}
\end{tabular}
where $e^{\text{RBF}}(\cdot): [0,\infty)\rightarrow \mathbb{R}^{d}$ is a distance expansion function to encode the distance variable into a $d$-dimensional vector \cite{torchmd} (see Appendix~\ref{appendix:method}). The terms $\mathbf{W}^{q}$, $\mathbf{W}^{k}$, $\mathbf{W}^{v}$, $\mathbf{W}^{dk}$, $\mathbf{W}^{dv}$, and $\mathbf{W}^{f}$ are learnable linear projectors, and $\odot$ denotes element-wise product. The output scalar features $\mathbf{o}_{i}^1$, $\mathbf{o}_{i}^2$, and $\mathbf{o}_{i}^3$ encode pairwise interactions and interatomic distances. Moreover, the attention mechanism facilitates the integration of distance information into the vector features via scalar filters $\mathbf{s}_{ij}^1$ and $\mathbf{s}_{ij}^2$ within the subsequent 3D Update Layer.

\textbf{3D Update Layer.} The 3D Update Layer facilitates information exchange between scalar and vector features. Vector features $\mathbf{vec}_i \in \mathbb{R}^{* \times 3}$ are initialized as zeros and jointly updated with scalar features $x_i$. The update $\Delta x_i$ and $\Delta \mathbf{vec}_i$ are defined as:
\vspace{-4mm}

\begin{tabular}{p{6.5cm}p{6.8cm}}
\begin{equation}
[\mathbf{u}_{i}^1, \mathbf{u}_{i}^2, \mathbf{u}_{i}^3]=\mathbf{W}^v\left(\mathbf{vec}_i\right), 
\end{equation}
&
\vspace{-8mm}
\begin{equation}
\mathbf{w}_j = \sum_{j=1}^N \left( \mathbf{vec}_j \odot \mathbf{s}_{ij}^1  \right)+ \mathbf{s}_{ij}^2 \odot \frac{\mathbf{r}_{i}-\mathbf{r}_{j}}{\left\| \mathbf{r}_i - \mathbf{r}_j \right\|},
\end{equation}
\\
\vspace{-10mm}
\begin{equation}
\Delta x_i = \mathbf{o}_{i}^2+\mathbf{o}_{i}^3 \odot (\mathbf{u}_{i}^1 \cdot \mathbf{u}_{i}^2), 
\end{equation}
&
\vspace{-10mm}
\begin{equation}
\Delta \mathbf{vec}_i = \mathbf{u}_{i}^3 \odot \mathbf{o}_{i}^1 + \mathbf{w}_{j},
\end{equation}
\end{tabular}
\vspace{-5mm}

where $\mathbf{W}^v$ are learnable linear projectors. Scalar features incorporate vector information via element-wise multiplication with the scalar product of vector components, while vector features are updated using both directional features $\mathbf{w}_j$ and a scalar filter $\mathbf{o}_i^1$.

As shown in Figure~\ref{fig:3dreltrans}, the 3D-ReTrans is constructed by stacking multiple 3D Relational-Attention and 3D Update Layers. Each layer performs residual updates on both scalar features $\Delta x_i$ and vector features $\Delta \mathbf{vec}_i$, allowing the model to simultaneously capture scalar properties (\eg atom types, distances) and expressive directional geometric information. For pretraining, the final outputs $x_i$ and $\mathbf{vec}_i$ are fed into the decoder, while for finetuning, they are passed to the prediction head.

\textbf{Learning 3D Equivariance and Invariance by Data Augmentations.} Considering that our model lacks built-in 3D equivariance or invariance, we leverage data augmentations to instill these symmetries, following AlphaFold3 \cite{AF3}. During MGM pretraining, atomic coordinates  $\mathbf{x}$ are randomly rotated using transformations sampled from the SO(3) group and translated with offsets drawn from  $\mathbf{t} \sim \mathcal{N}(\mathbf{0},0.01\mathbf{I}_3)$. These augmentations encourage the model to adjust its predictions equivariantly with any rotations and small translations. During fine-tuning for property prediction, the same augmentations are applied, but the model is trained to predict consistent properties, thereby learning invariance to rotations and translations.

We favor data-augmented equivariance with relational attention over fully E(3)-equivariant message passing. While built-in equivariant architectures offer formal guarantees, they often restrict how pair features are parameterized and incur non-trivial computational overhead (\eg tensor bases, spherical harmonics), which can hinder scaling and complicate integration with diverse molecular cues. In contrast, our approach instills rotational and translation robustness through augmentations, allowing encoder to operate with lightweight vector and scalar updates and to flexibly ingest high-dimensional pair features without architectural surgery. This yields a plug-and-play backbone that is easier to optimize, accommodates sparsity and density variations, and remains representation-rich: relational attention can expand or swap pair features as downstream tasks evolve, while maintaining competitive robustness to pose changes at a substantially lower training and inference cost.

%% file: sections/4_analysis.tex
\begin{figure}[t]
\centering
\begin{minipage}{0.33\textwidth}
    \centering
    \includegraphics[width=\textwidth]{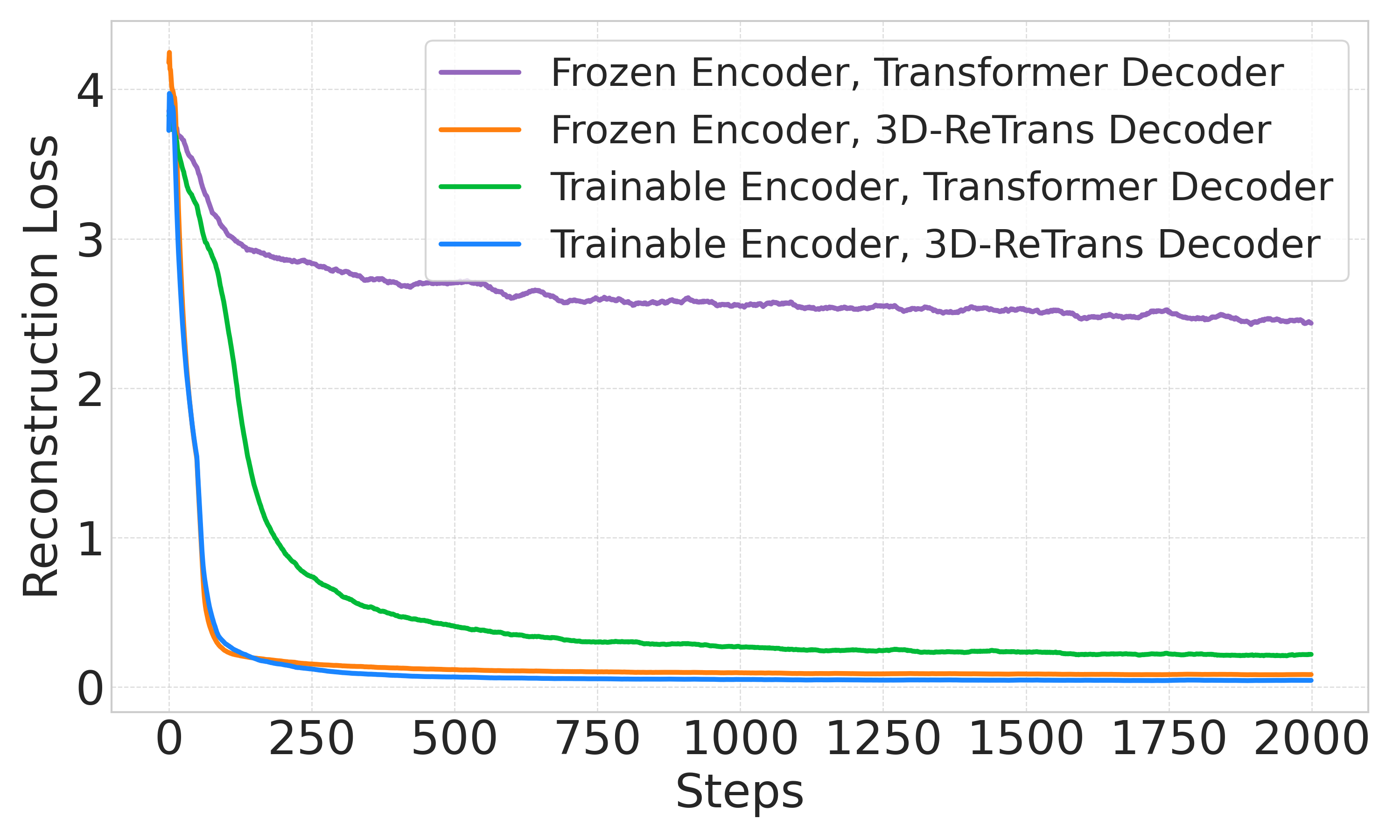}
    \vspace{-10pt}
    \caption{Reconstruction loss across four pretraining settings.}
    \label{fig:analysis1}
\end{minipage}
\hfill
\begin{minipage}{0.32\textwidth}
    \centering
    \includegraphics[width=\textwidth]{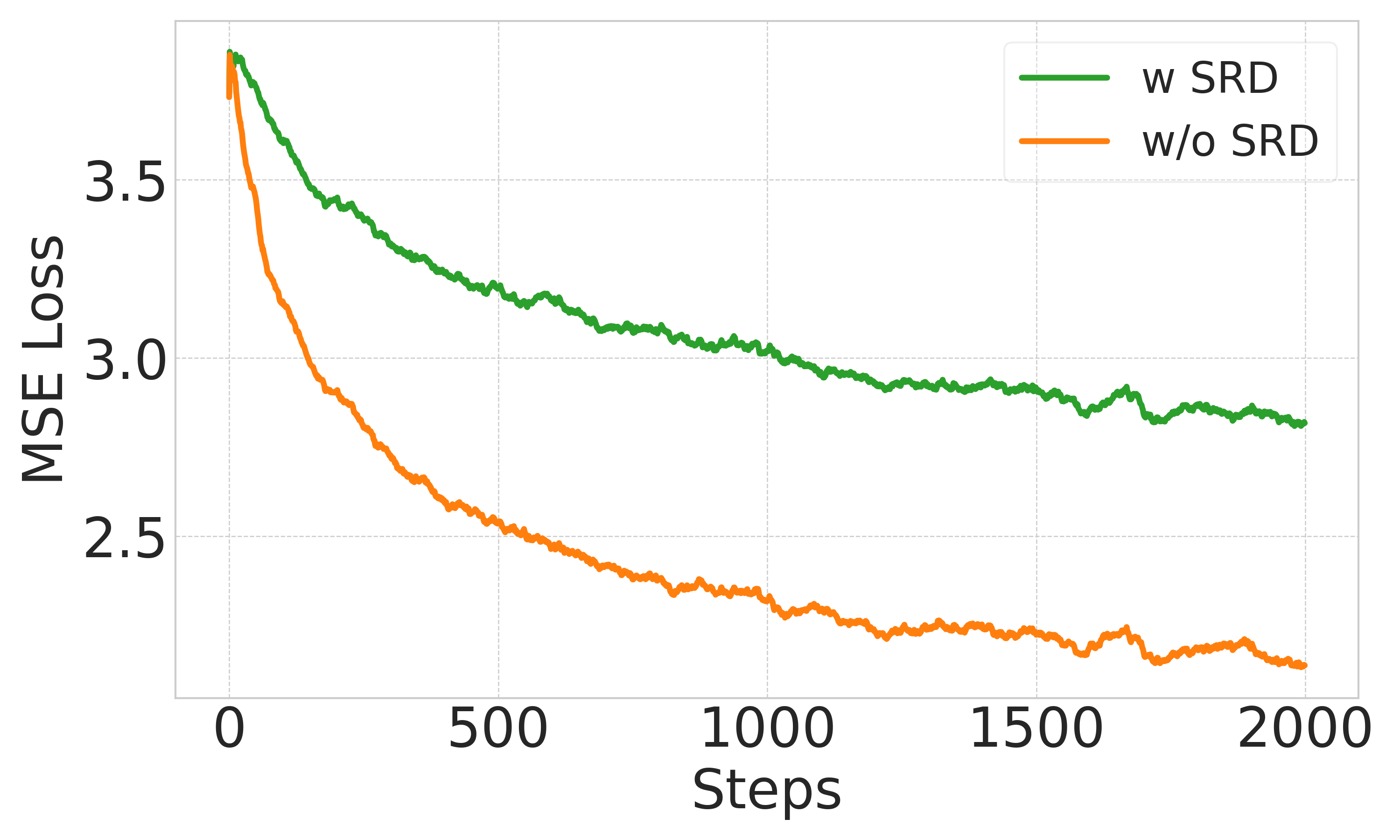}
    \vspace{-10pt}
    \caption{Probe encoder for masked atom coordinates when pretrained with/without SRD.}
    \label{fig:analysis4}
\end{minipage}
\hfill
\begin{minipage}{0.32\textwidth}
    \centering
    \includegraphics[width=\textwidth]{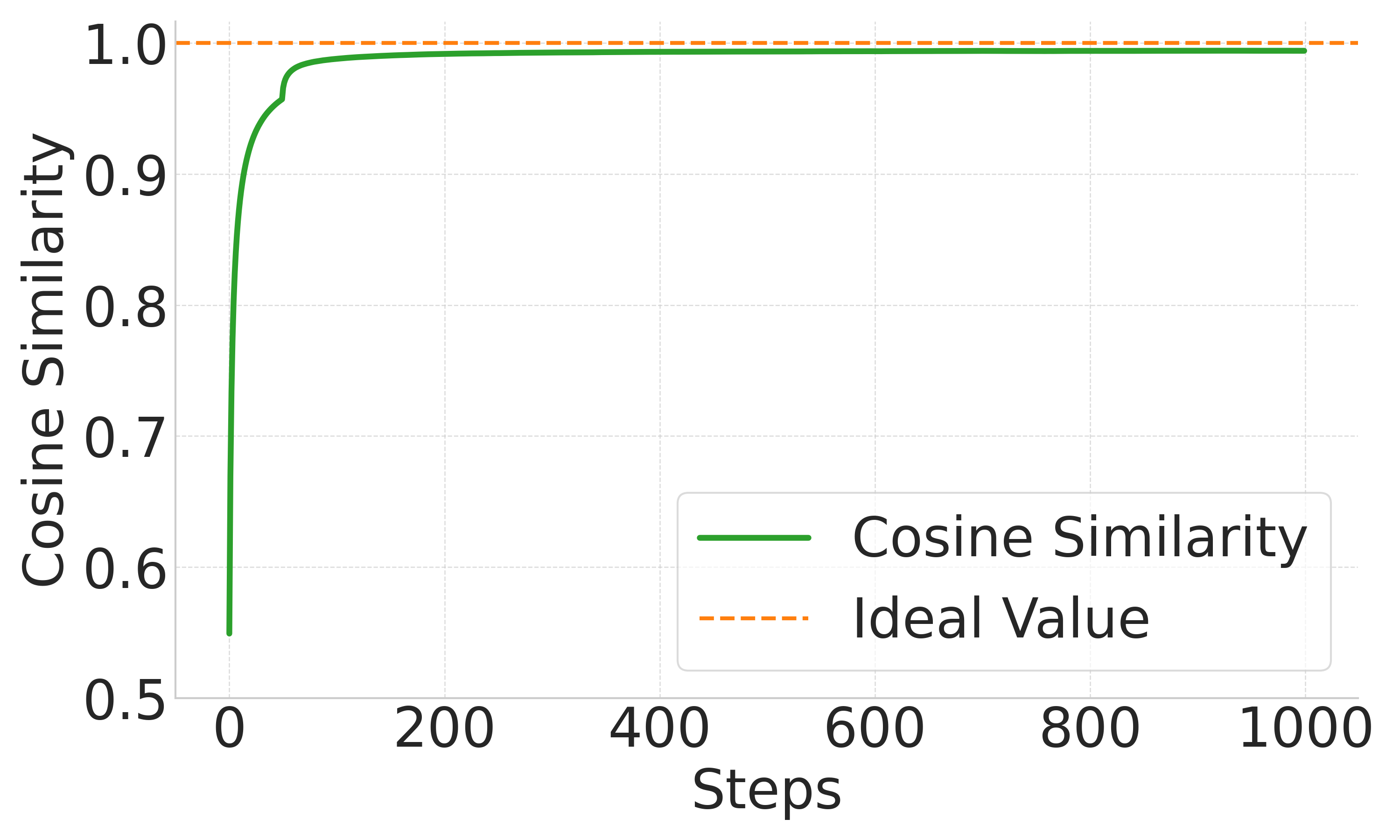}
    \vspace{-10pt}
    \caption{Reconstructing 2D-PE representation using 3D encoder representation.}
    \label{fig:analysis5}
\end{minipage}
\vspace{-15pt}
\end{figure}

\section{Analyzing Selective Re-mask Decoding and Structure-Independent Decoder}
\label{sec:analysis}

In this section, we conduct extensive experiments to evaluate the components of 3D-GSRD, focusing on the structure-independent decoder and SRD, including its key part 2D-from-3D distillation. We analyze their effects on the overall performance of the 3D-GSRD framework and how they contribute to 3D MGM pretraining.

\textbf{Analysis 1. The structure-dependent decoder can diminish the encoder's role in MRL.} We pretrain the auto-encoder framework under four settings, all using 3D-ReTrans as the encoder: (1) a frozen encoder with a Transformer decoder; (2) a frozen encoder with a 3D-ReTrans decoder; (3) a trainable encoder with a Transformer decoder; and (4) a trainable encoder with a 3D-ReTrans decoder. Figure~\ref{fig:analysis1} reports the reconstruction loss of masked atom coordinates, averaged over every 50 batches. When the encoder is frozen, the Transformer decoder (\ie structure-independent decoder) struggles to reconstruct masked atoms coordinates, yielding high reconstruction loss due to poor input representations and the absence of 2D molecular structural information.
In contrast, with the 3D-ReTrans decoder (\ie structure-dependent decoder), which leverages 2D molecular structures as input, the loss decreases rapidly during pretraining, even with a frozen encoder. This demonstrates that a powerful, structure-dependent decoder can compensate for weak encoder representations, diminishing the encoder's role in MRL.

\textbf{The structure-independent decoder heavily relies on high-quality encoder representations.} When paired with a trainable encoder, the structure-independent decoder achieves a much lower reconstruction loss, indicating that it relies heavily on the encoder to provide informative representations. In contrast, with a structure-dependent decoder, the loss remains low regardless of the encoder representation's quality, highlighting that such decoders reduce the learning pressure on the encoder and can hinder its ability to learn meaningful molecular features.


\textbf{Analysis 2. Structure-independent decoder improves downstream performance compared to structure-dependent decoder.} We pretrain the auto-encoder using either a structure-dependent or a structure-independent decoder while keeping all other settings constant, then finetune the pretrained encoder on molecular property prediction tasks. As Table~\ref{table:analysis} shows, the structure-independent decoder achieves better performance, outperforming the structure-dependent decoder by 5\% in Toluene. This demonstrates that using a structure-independent decoder encourages the encoder to learn more informative representations, leading to improved performance on downstream tasks.

\begin{wraptable}[11]{r}{0.49\textwidth}
\vspace{-3mm}
\centering
\caption{Analyzing the decoder and SRD. Performance (MAE $\downarrow$) on MD17. The variant without SRD and $\mathcal{L}_{\text{distill}}$ corresponds to the model ablated without 2D-PE. }
\vspace{-5pt}
\label{table:analysis}
\setlength{\tabcolsep}{2pt}
\begin{small}
\resizebox{0.48\textwidth}{!}{  
\begin{tabular}{lcccccc}
\toprule
Decoder & SRD & $\mathcal{L}_{\text{distill}}$ & Salicylic & Toluene & Uracil \\
\midrule
Structure-dependent & \ding{51} & \ding{51} & 0.0401 & 0.0291 &0.0334 \\
Structure-independent  & \cross  & \cross & 0.0404 & 0.0292 &0.0328 \\
Structure-independent  & \ding{51} & \cross  & 0.0416 &0.0293 & 0.0329 \\
Structure-independent  & \ding{51} & \ding{51} &\textbf{0.0387} & \textbf{0.0275} &\textbf{0.0315} \\
\bottomrule
\end{tabular}
}
\end{small}
\vspace{-2mm}
\end{wraptable}
\textbf{Analysis 3. 2D-PE and 2D-from-3D distillation boost downstream performance.} We pretrain the auto-encoder with and without 2D-PE and 2D-from-3D distillation, followed by finetuning the encoder on MD17 datasets. As shown in Table~\ref{table:analysis}, combining 2D-PE and distillation consistently improves performance. In contrast, using 2D-PE alone leads to degradation, likely due to unintended leakage of 2D structural information into the decoder. Moreover, 2D-from-3D distillation guides the 2D-PE to focus on encoding positional information for re-masked tokens, rather than learning molecular representation, which allows the 3D graph encoder to better specialize in MRL.

\textbf{Analysis 4. SRD prevents the encoder representation from containing information about 3D coordinates.} To examine whether the encoder captures detailed 3D coordinate information, we train an MLP probe to predict masked atom coordinates from the encoder’s representations. We compare the reconstruction loss for encoders pretrained with and without SRD. As shown in Figure~\ref{fig:analysis4}, the reconstruction loss for the encoder pretrained with SRD is much higher, suggesting that SRD suppresses direct encoding of 3D coordinate details. This forces the encoder to focus on learning higher-level molecular representations that are better aligned with downstream tasks.

\textbf{Analysis 5. 2D-PE produces 2D structural context without introducing information leakage.} 
To assess whether 2D-PE introduces additional information beyond what the 3D graph encoder already captures, we train an MLP to reconstruct the 2D-PE representation from the 3D graph encoder representation and compute the reconstruction error, measured by the cosine similarity between the two representations. During this process, both the 3D graph encoder and 2D-PE are frozen, and only the MLP is trainable. Figure~\ref{fig:analysis5} shows that the cosine similarity is very close to 1.0, indicating that the context 2D-PE provided is mostly contained by the 3D graph encoder.

\reb{\textbf{Analysis 6. 2D-PE encodes structural information for decoding.} We probe whether the pretrained 2D-PE captures structural information required by the decoder. Specifically, we freeze the 2D-PE and train two MLP classifiers to predict atom and bond types from its outputs. Both tasks achieve prediction accuracy above 99.99\%, demonstrating that the 2D-PE indeed encodes the structural information necessary for decoding.}

%% file: sections/5_experiments.tex
\section{Experiments}
\label{sec:experiments}

\subsection{Experimental Setup}
\label{sec:experimental Setup}

\textbf{Datasets.} For pretraining, we use a large-scale molecular dataset PCQM4Mv2 \cite{hu2020open}, which contains approximately 3.37 million equilibrium 3D molecular graph structures. For downstream tasks, we evaluate our model on two widely used molecular property prediction datasets: QM9 \cite{qm9} and MD17 \cite{md17}. Specifically, QM9 is a quantum chemistry dataset comprising 134k small molecules, each with its equilibrium conformation and 12 molecular properties (\eg homo, lumo, dipole moment \etc) calculated using density functional theory (DFT). Following prior works \cite{feng2023fractional, ni2023sliced}, we split the dataset into 11,000/1,000/10,831 molecules for training, validation, and testing, respectively. MD17 provides simulated dynamical trajectories for 8 small molecules, including their energy, forces, and conformations. During finetuning, our model first predicts the molecular energy and subsequently derives the forces using the relationship $F = -\nabla_r E$, where $r$ represents the 3D coordinates. For finetuning, we split the dataset into 9500/950 samples for training and validation, and use the remaining samples for testing.

\textbf{Baselines.} To evaluate the effectiveness of our proposed framework, we adopt state-of-the-art 3D molecular pretraining methods and supervised models for molecular property prediction as baselines. For 3D molecular pretraining methods, we include Transformer-M \cite{Transformer-M}, SE(3)-DDM \cite{se3ddm}, 3D-EMGP \cite{3d-emgp}, Coordinate Denoising \cite{zaidi2022pre}, Fractional Denoising \cite{feng2023fractional}, Sliced Denoising \cite{ni2023sliced}, UniGEM \cite{unigem}, Mol-AE \cite{molae}. For supervised models, we include SchNet \cite{schnet}, E(n)-GNN \cite{engnn}, DimeNet \cite{dimenet}, DimeNet++ \cite{dimenet++}, PaiNN \cite{painn}, SphereNet \cite{spherenet}, TorchMD-NET \cite{torchmd}, Uni-Mol2 \cite{unimol2}, ComENet \cite{comenet}. We also include the results of training our backbone (\ie 3D-ReTrans) from scratch to evaluate the effectiveness of our pretraining methods. We reproduce the results for Frad, SliDe and Mol-AE, while the results for other baselines are directly taken from the referenced papers. More details about baselines and implementation are provided in Appendix~\ref{appendix:setup}.

\begin{table*}[t]
\setlength{\tabcolsep}{3pt}
    \caption{Performance (MAE $\downarrow$) on MD17 force prediction. The best results are \textbf{bold}. The second-best results are \underline{underline}. Results marked with * are reproduced by us.}
    \vspace{-8pt}
    \label{table:md17}
    \vskip -0.15in
    \begin{center}
    \resizebox{\textwidth}{!}{
    \begin{footnotesize}
    \begin{tabular}{cccccccccc}
    \toprule
    Models & Aspirin	 & 	Benzene	 & 	Ethanol	 & 	Malonaldehyde	 & 	Naphthalene	 & 	Salicylic		 & Toluene		 & Uracil	\\ 
    \midrule
    TorchMD-NET  &  0.1216  & 0.1479  & 0.0492  & 0.0695  & \underline{0.0390} &  0.0655 &  \underline{0.0393} &  0.0484 \\
    3D-EMGP	 & 	0.1560		 & 0.1648	 & 	0.0389	 & 	0.0737	 & 	0.0829		 & 0.1187		 & 0.0619	 & 	0.0773\\
    \makecell[c]{3D-EMGP \\(TorchMD-NET)}	 & 	0.1124	 & 	\underline{0.1417}	 & 0.0445	& 0.0618 & 	0.0352	& 0.0586 & 0.0385 &0.0477\\
    Frad* &	0.0825 & \textbf{0.1355}	 &\underline{0.0432} & \underline{0.0535} & 0.0431 & 0.0569	&0.0433	 & 0.0482\\
    \rowcolor{lightblue} 
    3D-ReTrans &	\underline{0.0726} & 0.1619	 & 0.0556 &0.0659 & 0.0423 & \underline{0.0523}	&0.0417	 & \underline{0.0427}\\
    \midrule
    \rowcolor{lightblue} 
    3D-GSRD & \textbf{0.0583} & 0.1435 & \textbf{0.0355} & \textbf{0.0468} & \textbf{0.0266} & \textbf{0.0356} & \textbf{0.0274} & \textbf{0.0292}  \\
    \bottomrule
    \end{tabular}
    \end{footnotesize}
    }
    \end{center}
    \vskip -0.15in
\end{table*}

\begin{table*}[t]
\setlength{\tabcolsep}{3pt}
    \caption{Performance (MAE $\downarrow$) on QM9. The best results are \textbf{bold}. The second-best results are \underline{underline}. Results marked with * are reproduced by us. }
    \label{table:qm9}
    \vspace{-8pt}
    \begin{center}
    \resizebox{\textwidth}{!}{
    \begin{tabular}{lcccccccccccc}
    \toprule
    	Models & \makecell[c]{$\mu$ \\(D)}	& \makecell[c]{$\alpha$ \\($a_0^3$)}			&  \makecell[c]{$homo$ \\(meV)}		& \makecell[c]{$lumo$\\ (meV)}		& \makecell[c]{$gap$\\ (meV)}	& \makecell[c]{$<R^2>$ \\($a_0^2$)}	& \makecell[c]{ZPVE\\ (meV)	}	& \makecell[c]{$U_0$ \\ (meV)}		& \makecell[c]{$U$ \\ (meV)}		& \makecell[c]{$H$ \\ (meV)}		& \makecell[c]{$G$\\ (meV)} & \makecell[c]{$C_v$\\ ($\frac{cal}{mol K}$)	}
     \\
    \midrule
    \reb{Uni-Mol2} & 0.089 &0.305 &- &- &- &5.26 &- &- &- &- &- &0.144 \\
    SchNet & 	0.033 & 	0.235	 & 41.0 & 	34.0 & 63.0	 & 0.07	 & 1.70	 & 14.00	 & 19.00	 & 14.00	 & 14.00	 & 0.033\\
    E(n)-GNN & 	0.029	 & 0.071	 & 29.0	 & 25.0	 & 48.0	 & 0.11	 &   1.55	 & 11.00	 & 12.00	 & 12.00	 & 12.00	 & 0.031\\
    DimeNet++	 & 0.030	 & 0.043	 & 24.6	 & 19.5	 & 32.6	 & 0.33	 & 1.21	 & 6.32	 & 6.28 & 	6.53	 & 7.56	 & 0.023\\
    PaiNN	 & 0.012	 & 0.045	 & 27.6	 & 20.4	 & 45.7	 & 0.07	 & 1.28	 & \underline{5.85}	 & \underline{5.83}	 & \underline{5.98}	 & 7.35	 & 0.024\\
    SphereNet & 0.025 & 0.045 &22.8 &18.9 & 31.1  & 0.27 & \textbf{1.12} & 6.26 &  6.36 & 6.33 &7.78 &0.022\\ 
    \reb{ComENet} &0.025  &0.045 &23.1 &19.8 &32.4 &0.259 &\underline{1.20} &6.59 &6.82 & 6.86 &7.98 &0.024 \\ 
    TorchMD-NET & \underline{0.011} & 0.059 & 20.3 & 18.6 & 36.1 & \textbf{0.033} & 1.84 & 6.15 & 6.38 & 6.16 & 7.62 & 0.026 \\
   \rowcolor{lightblue} 
    3D-ReTrans & 0.016 &0.055 &22.0 & 17.8 &38.0 &0.341 &1.85 & 6.18 &6.36 &6.51 &7.89 &0.029 \\
   \midrule 
    Transformer-M &	0.037 &	\underline{0.041} &	\underline{17.5} &	16.2 &	\textbf{27.4} &	0.075 &		1.18 &	9.37 &	9.41 &	9.39 &	9.63 &	\underline{0.022}\\
     SE(3)-DDM 	&0.015	&0.046 &23.5  &19.5	 &40.2	&0.122	&1.31	&	6.92 &6.99 &7.09  &	7.65 &0.024 \\
    3D-EMGP &0.020	&0.057	&21.3	&18.2	&37.1  &0.092	&1.38 &	8.60 &8.60	&8.70 &	9.30  &0.026\\
    Coord  &0.016 &0.052 &17.7 &14.7 &31.8 &0.450 &1.71 &6.57 &6.11 &6.45 &\underline{6.91} &\textbf{0.020} \\
    Frad* &	0.012 &0.045 &\textbf{15.4} &\textbf{13.7} &30.6 &0.428 &1.56 &15.88 &14.67 & 14.87 &13.52 &0.023 \\
    SliDe* & 0.015 & 0.050 & 18.7 & 16.2 & \underline{28.8} & 0.606 & 1.78 & 10.05 & 10.79 & 11.34 & 11.80 & 0.025 \\
    \reb{Mol-AE*} &0.152 &0.434 &- &- &- &6.962 &- &- &- &- &- &0.215 \\
    \reb{Uni-GEM} &0.019 &0.060 &20.9 &16.7 &34.5  &- &- &- &- &-  &- &0.023 \\  
    \midrule   
   \rowcolor{lightblue} 
    3D-GSRD &\textbf{0.009} &\textbf{0.038} &18.0 &\underline{14.5} &31.1 &\underline{0.047} &1.38 &\textbf{5.48} &\textbf{5.67} &\textbf{5.84} &\textbf{6.90} &\textbf{0.020} \\ 
    \bottomrule
    \end{tabular}
    }
    \end{center}
    \vskip -0.2in
    \vspace{-5pt}
\end{table*}

\subsection{Results on MD17}
\label{sec:results on MD17}
The MD17 dataset contains diverse non-equilibrium molecular structures that are highly sensitive to geometry, making it a challenging benchmark for 3D MRL. As shown in Table~\ref{table:md17}, 3D-ReTrans achieves performance comparable to TorchMD-NET and surpasses 3D-EMGP on 7 of 8 molecules, demonstrating its strength as a 3D graph encoder for 3D MGM. Moreover, 3D-GSRD attains state-of-the-art results on 7 of 8 molecules except Benzene, exceeding the strongest baseline (\ie Frad) by a large margin. These results confirm the effectiveness of our pretraining method.

\subsection{Results on QM9}
\label{sec:results on QM9}

We also evaluate the effectiveness of 3D-ReTrans and our pretraining strategy on the QM9 dataset, as shown in Table~\ref{table:qm9}. 3D-ReTrans achieves performance comparable to TorchMD-NET, validating the effectiveness of our proposed backbone architecture. Moreover, 3D-GSRD sets a new state-of-the-art on 7 out of 12 properties, surpassing most baselines, including methods with and without pretraining. These results demonstrate that 3D-GSRD is a highly effective pretraining strategy for MRL, offering advantages over coordinate denoising based approaches.



\begin{table*}[t]
\centering
\begin{minipage}[t]{0.48\textwidth}
\centering
\caption{Ablation on 3D-ReTrans components. Performance (MAE $\downarrow$) on QM9.}
\label{tab:ablation on 3D-ReTrans components}
\resizebox{\linewidth}{!}{
\begin{tabular}{lccc}
\toprule
Model Components & homo & lumo & zpve \\
\midrule
Relational-Transformer & 27.7 & 24.0 & 1.97 \\
+ 3D Data Augmentation & 24.6 & 23.2 & 1.92 \\
+ 3D Relational-Attention & 23.4 & 20.3 & 1.90 \\
+ 3D Update Layer (3D-ReTrans) & \textbf{22.0} & \textbf{17.8} & \textbf{1.85} \\
\bottomrule
\end{tabular}
}
\end{minipage}\hfill
\begin{minipage}[t]{0.48\textwidth}
\centering
\caption{Analyzing SRD on the Relational-Transformer. Performance (MAE $\downarrow$) on MD17.}
\label{tab:relational-transformer}
\resizebox{\linewidth}{!}{
\begin{tabular}{llccc}
\toprule
Decoder & SRD & $\mathcal{L}_{\text{distill}}$ & Toluene & Uracil \\ 
\midrule
Structure-dependent & \ding{51} & \ding{51} & 0.1144 & 0.0813 \\
Structure-independent & \cross & \cross & 0.1250 & 0.0828 \\
Structure-independent & \ding{51} & \cross & 0.0998 & 0.0843 \\
Structure-independent & \ding{51} & \ding{51} & \textbf{0.0745} & \textbf{0.0733} \\
\bottomrule
\end{tabular}
}
\end{minipage}
\vspace{-10pt}
\end{table*}

\subsection{Ablation Studies and Analysis}
\label{sec:ablation}

\reb{\textbf{Ablation on Backbone.} To assess the effectiveness of our improvements to the Relational-Transformer~\cite{RelaTrans}, we perform ablation studies on each component of 3D-ReTrans, as summarized in Table~\ref{tab:ablation on 3D-ReTrans components}. The results show that 3D Relational-Attention, the 3D Update Layer, and 3D data augmentation each enhance molecular property prediction, collectively boosting overall performance.}

\reb{\textbf{Generalization of SRD Across 3D Graph Encoders.} To evaluate the generalization of SRD, we replace 3D-ReTrans with the Relational-Transformer and conduct additional experiments. As shown in Table~\ref{tab:relational-transformer}, incorporating SRD consistently improves  downstream performance, demonstrating its effectiveness as a general pretraining strategy applicable to diverse 3D graph encoder architectures.}

\begin{wraptable}{r}{0.35\textwidth}
\vspace{-3mm}
\centering
\caption{Ablation on 2D graph position encoder. Performance (MAE $\downarrow$) on MD17.}
\vspace{-1mm}
\label{tab:Ablation on 2D encoder}
\setlength{\tabcolsep}{3pt}
\begin{small}
\begin{tabular}{lcc}
\toprule
2D Encoder & Salicylic & Uracil \\
\midrule
RWSE & 0.0368 & 0.0310 \\
2D-PE & \textbf{0.0356} & \textbf{0.0292} \\
\bottomrule
\end{tabular}
\end{small}
\end{wraptable}

\textbf{Ablation on 2D Graph Position Encoder.} We compare our 2D-PE against alternative structural embeddings, such as those in GraphGPS \cite{GraphGPS}. Following prior results, we adopt RWSE \cite{dwivedi2021graph} as a representative baseline due to its strong performance on ZINC and PCQM4Mv2 with relatively low computational cost. 
As shown in Table~\ref{tab:Ablation on 2D encoder}, replacing 2D-PE with RWSE leads to consistently lower performance, confirming the advantage of 2D-PE in providing 2D structural context.

%% file: sections/6_conclusion.tex
\section{Conclusion and Future Works}
\label{sec:conclusion}
In this work, we introduce 3D-GSRD, a 3D MGM framework with three key components: (1) the Selective Re-mask Decoding that selectively re-masks 3D-relevant information while preserving 2D graph structures; (2) a structure-independent decoder that eliminates all structural information by relying solely on encoder representation; and (3) 3D-ReTrans as the 3D graph encoder for MRL. Our detailed analysis reveals the internal mechanisms of SRD and the structure-independent decoder. Extensive experiments demonstrate that 3D-GSRD significantly outperforms baselines on downstream datasets such as QM9 and MD17. 

Despite promising results, several limitations remain. Our pretraining is conducted on PCQM4Mv2 \cite{hu2020open} with 3.37M molecules, which is smaller than large-scale datasets such as PubChemQC \cite{pubchemqc} with 230M molecules, potentially constraining performance. Scaling to larger and more diverse datasets is an important direction. In addition, we focus on molecular property prediction, while other tasks like 3D molecule generation \cite{liu2025nextmol,luo2025towards,wang2025survey,xu2025cofm} and multi-modal molecule–text modeling \cite{liu2023molca,3dmolm,liu2024reactxt,chen2025hierarchical} could also benefit from our pretrained autoencoder. Beyond molecular applications, our pretraining paradigm can be extended to broader biological modalities such as single-cell \cite{shi2025language} and protein \cite{liu2024prott}.

\section*{Acknowledgement}

This research is supported by the National Natural Science Foundation of China (62572449). 

%% file: sections/8_appendix.tex


\section{Broader Impacts}
\label{appendix:Broader Impacts}

This work advances molecular representation learning and has the potential to accelerate downstream applications such as molecular property prediction, drug discovery, molecular dynamics simulation, and material design, helping reduce the cost and time of wet-lab experiments. However, there is a risk of over-reliance on model predictions without sufficient interpretability or domain validation. Additionally, models trained on biased datasets may produce structural or chemical biases, limiting the model's generalization across molecular spaces. We encourage the community to test our models strictly before applying them in scientific scenarios.

\section{More Details on Methodology}
\label{appendix:method}
In this section, we describe the embedding layer preceding the attention mechanism in 3D-ReTrans. The input 3D molecule graph is defined as $G = (\mathbf{x},\mathbf{a},\mathbf{e})$, where $\mathbf{x}$ denotes atomic coordinates, $\mathbf{a}$ represents atom types, and $\mathbf{e}$ denotes pairwise edge features. Our goal is to obtain atomic and edge embeddings that encode both chemical and geometric context.

The initial node embedding $e^{\text{node}}$ jointly encodes atom coordinates and types:
\begin{equation}
    e^{\text{node}} = \text{Embed}^{\text{node}}([\mathbf{x}, \mathbf{a}]).
\end{equation}

To incorporate local geometric context, we compute the neighborhood embedding $e_{i}^{\text{neigh}}$ for each atom $i$ based on the radial distances $d_{ij}$ to neighbors atom $j$. The radial basis function is given by:
\begin{equation}
e^{\mathrm{RBF}}(d_{ij}) = \phi(d_{ij}) \exp\left(-\beta \left( \exp(-d_{ij}) - \mu \right)^2 \right)
\end{equation}
\begin{equation}
\phi(d_{ij}) = 
\begin{cases}
\frac{1}{2} \left( \cos\left( \frac{\pi d_{ij}}{d_{\mathrm{cut}}} \right) + 1 \right), & \text{if } d_{ij} \leq d_{\mathrm{cut}} \\
0, & \text{if } d_{ij} > d_{\mathrm{cut}}
\end{cases}
\end{equation}
where $\phi(d_{ij})$ is a smooth cutoff function ensuring locality and $\beta$ and $\mu$ are fixed parameters.

The neighborhood embedding $e_{i}^{\text{neigh}}$ for atom $i$ is then defined as:
\begin{align}
e_{i}^{\text{neigh}} &= \sum_{j=1}^N\text{Embed}^{\text{neigh}}([\mathbf{x}, \mathbf{a}]) \odot \mathbf{W}^{r} e^{\mathrm{RBF}}(d_{ij}),
\end{align}
where $\odot$ denotes element-wise product and $\mathbf{W}^{r}$ is a learnable projector. We set the cutoff distance $d_{\mathrm{cut}}=5\,\text{\AA}$, ensuring that each atom only attends to neighbors within this spatial range. 

The final atomic embedding $e_i^{\text{atomic}}$ combines node and neighborhood information:
\begin{align}
e_i^{\text{atomic}} &= \mathbf{W}^{a}([e^{\text{node}},e_{i}^{\text{neigh}}]),
\end{align}
where $\mathbf{W}^{a}$ denotes learnable projector.

We also obtain the edge embedding $e^{\text{edge}}$ via:
\begin{align}
e^{\text{edge}} &= \text{Embed}^{\text{edge}}(\mathbf{e}).
\end{align}

\section{Pseudo Code}
We present the pseudocode for pretraining (see Algorithm~\ref{alg:pretraining}) and finetuning (see Algorithm~\ref{alg:finetuning}) algorithms in this section.

\begin{algorithm}[htbp]
\caption{Pretraining of 3D-GSRD}\label{alg:pretraining}
\begin{algorithmic}[1]
\Require 3D graph encoder $\phi^{3D}_{\theta}$, 2D graph encoder $\phi^{2D}_{\theta}$, decoder $\phi^{De}_{\theta}$, pretraining dataset $D$, input 3D molecule graph $G = (\mathbf{x},\mathbf{a},\mathbf{e})$, masked coordinates prediction head $PosHead_{\theta}$, denoising prediction head $DenoiseHead_{\theta}$, mask ratio $p$, denoising loss weight $w$.

\While{training is not finished}
    \State $G_i = (\mathbf{x},\mathbf{a},\mathbf{e})$ = dataloader($D$) 
    \State randomly mask $p$ atoms and add Gaussian noise ($\Delta{x_i} \sim 0.04 \cdot \mathcal{N}(0, {\sigma}^2I_{m}) $) to the unmasked atomic coordinates
    \State input moleule $\tilde{G_i}=(\tilde{\mathbf{a}},\tilde{\mathbf{x}},\tilde{\mathbf{e}})$
    \State $\mathbf{h_{3D}},\mathbf{vec} =\phi^{3D}_{\theta}(\tilde{G_i})$ 
    \State $\mathbf{h_{2D}}=\phi^{2D}_{\theta}(\mathbf{a},\mathbf{e}))$ 
    \State $\text{SRD}(\mathbf{h_{3D}}, \tilde{G_i})=\text{re-mask}(\mathbf{h_{3D}}) + \text{stop-grad}(\mathbf{h_{2D}})$
    \State  $\mathbf{rep},\mathbf{vec} = \phi^{De}_{\theta}(\text{SRD}(\mathbf{h_{3D}}, \tilde{G_i}),\mathbf{vec})$ 
    \State For masked atoms: ${x_i^{pred}} = PosHead_{\theta}(\mathbf{rep},\mathbf{vec})$
    \State For unmasked atoms: $\Delta{x_i^{pred}}= DenoiseHead_{\theta}(\mathbf{rep},\mathbf{vec})$
    \State Loss = $||{x_i}^{pred} - {x_i}||_{2}^{2} +w \cdot ||\Delta{x_i}^{pred} - \Delta{x_i}||_{2}^{2} -CosineSimilarity(\text{stop-grad}(\mathbf{h_{3D}}),\mathbf{h_{2D}}) $ 
    \State Optimise(Loss)
\EndWhile
\end{algorithmic}
\end{algorithm}

\begin{algorithm}[htbp]
\caption{Finetuning of 3D-GSRD}\label{alg:finetuning}
\begin{algorithmic}[1]
\Require 3D graph encoder $\phi^{3D}_{\theta}$, finetuning dataset $D$, input 3D molecule graph $G = (\mathbf{x},\mathbf{a},\mathbf{e})$, label prediction head $LabelHead_{\theta}$.

\While{training is not finished}
    \State $G_i = (\mathbf{x},\mathbf{a},\mathbf{e}), y_i$ = dataloader($D$) 
    \State $\mathbf{h_{3D}},\mathbf{vec}=\phi^{3D}_{\theta}(G_i)$ 
    \State ${y_i^{pred}} = LabelHead_{\theta}(\mathbf{h_{3D}},\mathbf{vec})$
    \State Loss = $||{y_i}^{pred} - {y_i}||_{2}^{2}$ 
    \State Optimise(Loss)
\EndWhile
\end{algorithmic}
\end{algorithm}

\section{Experimental Setup}
\label{appendix:setup}

\subsection{Computational Resource}
All experiments are conducted on NVIDIA A6000-48G GPUs. Pretraining requires a total of 48 GPU hours. For downstream tasks, finetuning on QM9 and MD17 takes approximately 48 and 8 GPU hours per experiment, respectively.

\subsection{Baselines}
\label{appendix:baselines}
We describe the details of our reported baseline methods in this section.

\textbf{SchNet} \cite{schnet} proposes continuous-filter convolutional layers, which enables the model to capture local correlations in molecules without grid-based data.

\textbf{E(n)-GNN} \cite{engnn} introduces an architecture that is equivariant to rotations, translations, reflections, and permutations. Notably, its equivariance extends to higher dimensions with affordable computational overhead increase.

\textbf{DimeNet} \cite{dimenet} applies directional message passing which enables graph neural networks to incorporate directional information for molecular predictions and using spherical Bessel functions and spherical harmonics to representation distances and angles.

\textbf{DimeNet++} \cite{dimenet++} improves upon DimeNet by being 8× faster and achieving 10\% higher accuracy, while maintaining strong generalization across molecular configurations and compositions.

\textbf{PaiNN} \cite{painn} addresses the limitations of invariant representations in message passing neural networks by extending the message passing framework to rotationally equivariant representations.

\textbf{SphereNet} \cite{spherenet} analyzes 3D molecular graphs in the spherical coordinate system and propose the spherical message passing (SMP) scheme to efficiently distinguish molecular structures while reducing training complexity.

\textbf{TorchMD-NET} \cite{torchmd} introduces an equivariant Transformer architecture with a modified attention mechanism that incorporates interatomic distances directly into the attention weights.

\textbf{Transformer-M} \cite{Transformer-M} is a Transformer based architecture that that handles multiple molecular data modalities within a unified model by using two separate channels to encode 2D and 3D molecular structures.

\textbf{SE(3)-DDM} \cite{se3ddm} leverages an SE(3)-invariant score matching method to transform coordinate denoising into denoising the pairwise atomic distances within a molecule.

\textbf{3D-EMGP} \cite{3d-emgp} introduces an equivariant energy-based model and develops a self-supervised pretraining framework including a physics-inspired node-level force prediction task and a graph-level noise scale prediction task.

\textbf{Coord} \cite{zaidi2022pre} proposes a pretraining technique based on denoising for 3D molecular structures, showing it is equivalent to learning a force field.

\textbf{Frad} \cite{feng2023fractional} introduces a new hybrid noise strategy by first adding Gaussian noise to the dihedral angles of the rotatable bonds, followed by traditional noise to the atom coordinates, with pretraining focused solely on denoising the latter.

\textbf{SliDe} \cite{ni2023sliced} develops a novel sliced denoising method that adds Gaussian noise to bond lengths, angles, and torsion angles with their variances determined by parameters within the energy function.

\textbf{Uni-Mol2} \cite{unimol2} is a molecular pretraining model that uses a two-track transformer to jointly capture atomic-level, graph-level, and geometry-level features, while systematically investigating scaling laws in molecular pretraining.

\textbf{ComENet} \cite{comenet} introduces a graph neural network for 3D molecular graphs that adopts rotation angles and local completeness in the 1-hop neighborhood, while integrating quantum-inspired basis functions into its message passing mechanism.

\textbf{Mol-AE} \cite{molae} addresses the gap between pretraining and downstream objectives in encoder-only 3D molecular models by introducing an auto-encoder with positional encodings as atomic identifiers and a 3D Cloze Test objective that drops atoms to better capture real substructures.

\textbf{Uni-GEM} \cite{unigem} unifies molecular generation and property prediction through a diffusion-based two-phase process of scaffold nucleation and molecule growth, using a multi-branch network with oversampling to balance tasks.

\subsection{Implementation details}
\label{appendix:details}

We employ the 3D-ReTrans as the 3D graph encoder and implement 2D-PE as 2D-ReTrans, a simplified version of the 3D-ReTrans that excludes 3D coordinates and distance inputs, while using the Transformer as the structure-independent decoder. The 3D graph encoder is configured with a hidden dimension of 256, 8 attention heads, and 12 layers. The 2D-PE shares most of its configuration with the 3D graph encoder, except for a hidden dimension of 64 and 4 attention heads. The decoder consists of 2 layers, with the hidden dimension and number of attention heads same as the 3D graph encoder. The detailed hyper-parameters configuration for pretraining and finetuning are shown in Table~\ref{tab:pcqm4mv2}, Table~\ref{tab:qm9}, and Table~\ref{tab:md17}, respectively.

\begin{table*}[t]
    \caption{Hyper-parameters for pretraining on PCQM4MV2. }
    \label{tab:pcqm4mv2}
    \vskip 0.15in
    \begin{center}
    \begin{small}
    \begin{tabular}{lc}
    \toprule
    Parameter & Value\\
     \midrule
   Dataset & PCQM4MV2	\\
   Train/Val/Test Split &	Others/100/100 \\
   Batch size & 	128	\\ 
   Inference Batch size & 	128	\\ 
   Accumulate grad batches & 2 \\
   Optimizer  & 	AdamW	\\
   Weight decay & 	1e-16	\\
   Scheduler & CosineAnnealingLR \\
   Init learning rate & 5e-5 \\
   Min learning rate & 1e-6 \\
   Warm up steps & 10000	\\
   Max epochs & 30 \\
   Masked ratio & 0.25 \\
   Masked coordinates reconstruction loss type & MSE loss \\
   Coordinate noise scale(type: Gaussian) & 	0.04	\\
   Denoising loss weight & 0.1	\\
   \bottomrule
    \end{tabular}
    \end{small}
    \end{center}
    \vskip -0.2in
\end{table*}

\begin{table}[t]
    \caption{Hyper-parameters for finetuning on QM9. }
     \label{tab:qm9}
     \vskip 0.15in
    \begin{center}
    \begin{small}
    \begin{tabular}{lc}
    \toprule
    Parameter & Value\\
     \midrule
  Dataset & QM9 \\
  Train/Val/Test Split &	11000/1000/10831 \\
  Batch size &	128	\\
Inference Batch size & 	128	\\ 
  Accumulate grad batches & 1 \\
  Optimizer&	AdamW	\\
  Weight decay & 	1e-16	\\
  Scheduler & CosineAnnealingLR \\
  Init learning rate & 5e-4 \\
  Min learning rate & 1e-6 \\
  Warm up steps	&1000	\\
  Learning rate cosine length & 2,000,000 \\ 
  Max steps & 2,000,000 \\
  Max epochs & 2000 \\
  Finetuning loss type & MSE loss \\
   \bottomrule
    \end{tabular}
    \end{small}
    \end{center}
\end{table}

\begin{table}[t]
    \caption{Hyper-parameters for finetuning on MD17. }
     \label{tab:md17}
     \vskip 0.15in
    \begin{center}
    \begin{small}
    \begin{tabular}{lc}
    \toprule
    Parameter & Value\\
     \midrule
  Dataset & MD17 \\
  Train/Val Split &	9500/500/remaining data	\\
  Batch size &	80	\\ 
  Inference batch size &	64	\\ 
  Accumulate grad batches & 1 \\
  Optimizer&	AdamW	\\
  Weight decay & 	0.0	\\
  Scheduler & CosineAnnealingLR \\
  Init learning rate & 5e-4 \\
  Min learning rate & 1e-6 \\
  Warm up steps	&1000	\\	  
  Max epochs & 1200 \\
  Force weight	&0.8		\\
  Energy weight	&0.2		\\
  Finetuning loss type & MAE loss \\
  Ema alpha dy & 1.0 \\
  Ema alpha y & 0.05 \\
   \bottomrule
    \end{tabular}
    \end{small}
    \end{center}
\end{table}

\section{More Experimental Results}

\subsection{Ablation on Alternative Approach to Eliminating 2D Leakage}

\reb{We investigate an alternative approach to fully eliminate 2D leakage by directly combining 3D embeddings with 2D positional encodings as molecular representations during downstream tasks. We test this setup with SRD both with and without distillation. As shown in Table~\ref{table:alt_2d_leakage}, simply fusing 2D-PE with 3D-ReTrans under SRD without distillation consistently underperforms our original design, highlighting the necessity of SRD with distillation for effective 2D and 3D alignment. In contrast, when distillation is applied, downstream performance becomes insensitive to whether 2D-PE is explicitly included, suggesting that pretraining distillation sufficiently aligns the modalities and renders additional 2D input unnecessary.}

\subsection{Evolution of 2D-PE's representations}
\begin{figure}
    \centering
    \includegraphics[width=0.5\textwidth]{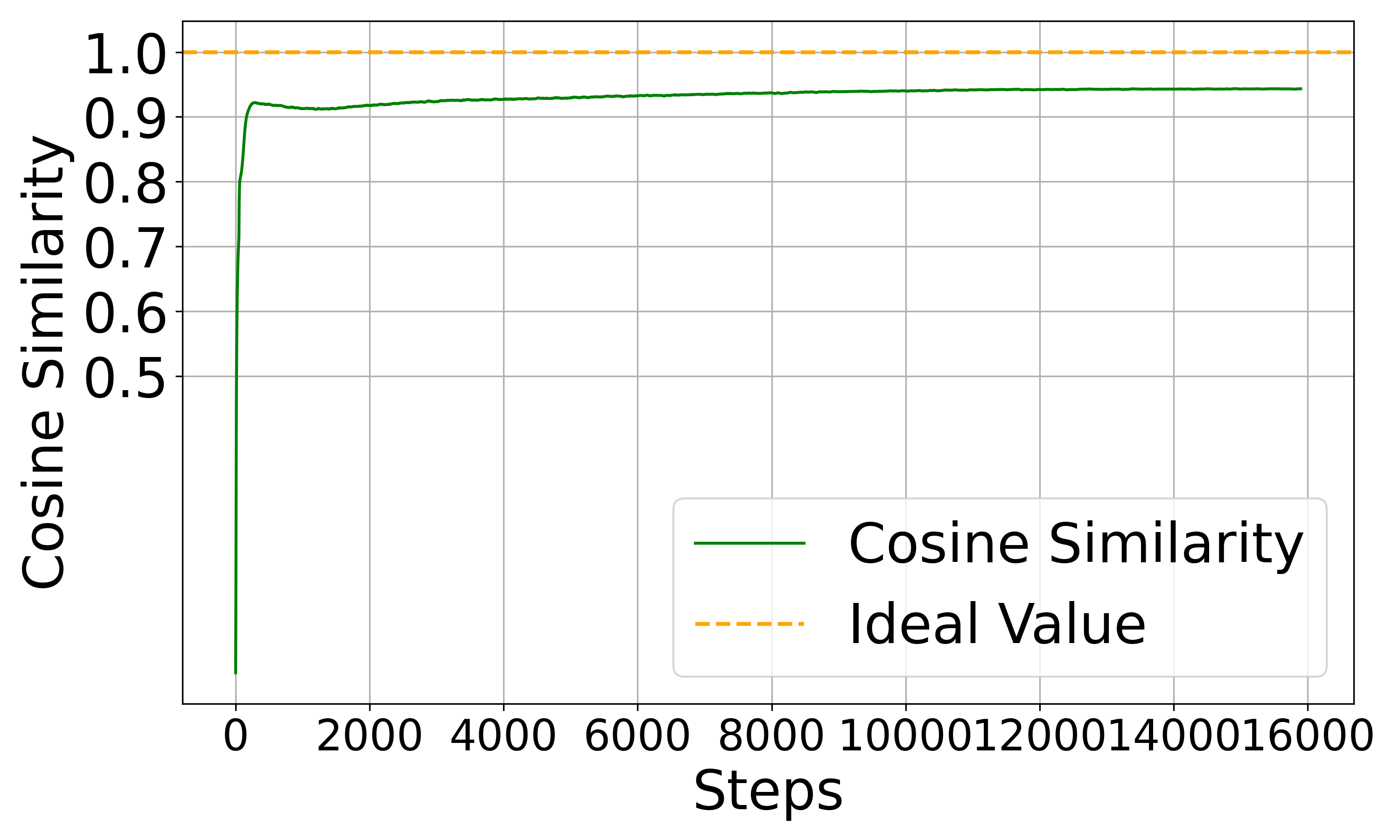}
    \caption{Evolution of the 2D-PE’s representation.}
    \label{fig:analysis7}
\end{figure}

To examine how 2D-PE's representation evolves during pretraining, we track the cosine similarity between 2D and 3D representations. As shown in Figure~\ref{fig:analysis7}, the similarity increases sharply in the initial training steps, approaching 1.0, and then grows gradually, indicating progressive alignment between the two modalities.

\begin{table}[t]
\vspace{-3mm}
\centering
\caption{Ablation on alternative approach to eliminate 2D leakage. Performance (MAE $\downarrow$) on MD17.}
\vspace{5pt}
\label{table:alt_2d_leakage}
\setlength{\tabcolsep}{3pt}
\resizebox{0.6\textwidth}{!}{  
\begin{tabular}{lcccccc}
\toprule
Downstream Model & SRD & $\mathcal{L}_{\text{distill}}$ & Salicylic & Toluene & Uracil \\
\midrule
2D-PE+3D-ReTrans & \ding{51} & \cross & 0.0420 & 0.0293 & 0.0334 \\
3D-ReTrans       & \ding{51} & \cross & 0.0416 & 0.0293 & 0.0329 \\
2D-PE+3D-ReTrans & \ding{51} & \ding{51} & 0.0384 & 0.0275 & 0.0311 \\
3D-ReTrans       & \ding{51} & \ding{51} & 0.0387 & 0.0275 & 0.0315 \\
\bottomrule
\end{tabular}
}
\vspace{-2mm}
\end{table}

